\documentclass[final]{siamltex}

\usepackage{amsmath,amsfonts, graphicx}
\pdfoutput=1

\usepackage{amsmath} 
\usepackage{amssymb}  

\usepackage{amsfonts}
\usepackage{cite}

\usepackage[usenames]{color}

\def\crossout#1{\st{#1}}

\def\qmark#1{{\color{green}?}\ul{#1}{\color{green}?}}

\def\coloremph#1{{\color{red}{#1}}}
\def\crossoutg#1{\colorbox{red}{\tiny DELETE START}#1\colorbox{red}{\tiny DELETE END}}
\def\crossout#1{}
\def\qmark#1{}
\def\coloremph#1{}
\def\crossoutg#1{}

\def\mindex#1{\index{#1}}

\def\sq{\hbox{\rlap{$\sqcap$}$\sqcup$}}
\def\qed{\ifmmode\sq\else{\unskip\nobreak\hfil
\penalty50\hskip1em\null\nobreak\hfil\sq
\parfillskip=0pt\finalhyphendemerits=0\endgraf}\fi\medskip}

\long\def\defbox#1{\framebox[.9\hsize][c]{\parbox{.85\hsize}{%
\parindent=0pt
\baselineskip=12pt plus .1pt      
\parskip=6pt plus 1.5pt minus 1pt 
 #1}}}

\long\def\beginbox#1\endbox{\subsection*{}%
\hbox{\hspace{.05\hsize}\defbox{\medskip#1\bigskip}}%
\subsection*{}}

\def\endbox{}

\def\transpose{{\hbox{\it\tiny T}}}

\newsavebox{\junk}
\savebox{\junk}[1.6mm]{\hbox{$|\!|\!|$}}

\def\limsup{\mathop{\rm lim\ sup}}

\def\argmin{\mathop{\rm arg\, min}}

\def\U{{\sf U}}

\def\state{{\sf X}}

\def\bx{{{\cal B}(\state)}}

\newcommand{\field}[1]{\mathbb{#1}}

\def\Re{\field{R}}

\def\ind{\field{I}}

\def\bfmath#1{{\mathchoice{\mbox{\boldmath$#1$}}%
{\mbox{\boldmath$#1$}}%
{\mbox{\boldmath$\scriptstyle#1$}}%
{\mbox{\boldmath$\scriptscriptstyle#1$}}}}

\def\bfmx{\bfmath{x}}

\def\bfmu{\bfmath{u}}

\def\bfmA{\bfmath{A}}
\def\bfmB{\bfmath{B}}

\def\bfmG{\bfmath{G}}

\def\bfmN{\bfmath{N}}

\def\bfmU{\bfmath{U}}

\def\bfmX{\bfmath{X}}

\def\bfmY{\bfmath{Y}}

\def\bfmhhaY{\bfmath{\hhaY}} 
\def\bfmhhaY{\hbox to 0pt{$\widehat{\bfmY}$\hss}\widehat{\phantom{\raise 1.25pt\hbox{$\bfmY$}}}}

\def\bfmW{\bfmath{W}}  
  
\def\bfmZ{\bfmath{Z}}

\def\til={{\widetilde =}}

\def\clD{{\cal D}}

\def\clF{{\cal F}}

\def\clP{{\cal P}}

 \def\FRAC#1#2#3{\genfrac{}{}{}{#1}{#2}{#3}}

\def\ddt{{\mathchoice{\FRAC{1}{d}{dt}}%
{\FRAC{1}{d}{dt}}%
{\FRAC{3}{d}{dt}}%
{\FRAC{3}{d}{dt}}}}

\def\ddtp{{\mathchoice{\FRAC{1}{d^{\hbox to 2pt{\rm\tiny +\hss}}}{dt}}%
{\FRAC{1}{d^{\hbox to 2pt{\rm\tiny +\hss}}}{dt}}%
{\FRAC{3}{d^{\hbox to 2pt{\rm\tiny +\hss}}}{dt}}%
{\FRAC{3}{d^{\hbox to 2pt{\rm\tiny +\hss}}}{dt}}}}

\def\half{{\mathchoice{\FRAC{1}{1}{2}}%
{\FRAC{1}{1}{2}}%
{\FRAC{3}{1}{2}}%
{\FRAC{3}{1}{2}}}}

\def\third{{\mathchoice{\FRAC{1}{1}{3}}%
{\FRAC{1}{1}{3}}%
{\FRAC{3}{1}{3}}%
{\FRAC{3}{1}{3}}}}

\def\eqdef{\mathbin{:=}}

\def\Prob{{\sf P}}

\def\Expect{{\sf E}}

\def\average#1,#2,{{1\over #2} \sum_{#1}^{#2}}

\def\eye(#1){{\bf(#1)}\quad}
\def\epsy{\varepsilon}

\def\varble{\,\cdot\,}

\def\Lemma#1{Lemma~\ref{#1}}
\def\Proposition#1{Proposition~\ref{#1}}

\def\Section#1{Section~\ref{#1}}

\def\Figure#1{Figure~\ref{#1}}

\def\eq#1/{(\ref{e:#1})}

\newcommand{\beqn}[1]{\notes{#1}%
\begin{eqnarray} \elabel{#1}}

\newcommand{\eeqn}{\end{eqnarray} }

\newcommand{\beq}[1]{\notes{#1}%
\begin{equation}\elabel{#1}}

\newcommand{\eeq}{\end{equation}}

\def\bdes{\begin{description}}
\def\edes{\end{description}}

\def\barf{{\overline {f}}}

\newcounter{rmnumsiam}

\newcounter{anum}

{\end{list}}

\def\ass(#1:#2){(#1\ref{#1:#2})}

\def\ritem#1{
\item[{\sf \ass(\current_model:#1)}]
}

\newenvironment{recall-ass}[1]{%
\begin{description}
\def\current_model{#1}}{
\end{description}
}

\def\Ebox#1#2{%
\begin{center}
 \parbox{#1\hsize}{\epsfxsize=\hsize \epsfbox{#2}}
\end{center}}

\newcommand{\bd}{\begin{description}}
\newcommand{\ed}{\end{description}}
\newcommand{\bt}{\begin{theorem}}
\newcommand{\et}{\end{theorem}}
\newcommand{\ba}{\begin{array}{rcl}}
\newcommand{\ea}{\end{array}}

\def\Lemma#1{Lemma~\ref{t:#1}}
\def\Proposition#1{Proposition~\ref{t:#1}}

 \def\Figure#1{Fig.~\ref{#1}}
 \def\Section#1{Sec.~\ref{#1}}

\def\tilx{\tilde{x}}
\def\tilbeta{\tilde{\beta}}

\def\td{{\hbox{\tiny TD}}}

\def\Ebox#1#2{%
\begin{center}
\includegraphics[width=#1\hsize]{./figures/#2}
\end{center}}

\def\Ecost{{\cal E}_{\hbox{\tiny c}}}

\def\tF{{\hbox{\tiny \rm F}}}
\def\tD{{\hbox{\tiny \rm D}}}

\def\generate{\clD} 
\def\generateF{\clD^\tF} 
\def\generateD{\clD^\tD} 

 \def\phiF{{\phi^{\tF*}}}
 \def\tJ{\hbox{\tiny \rm J}}
 \def\tK{\hbox{\tiny \rm K}}
 \def\phiJstar{{\phi^{\tJ*}}}

\def\barX{{\overline {X}}}

 \def\tK{\hbox{\tiny \rm K}}
 \def\phiKstar{{\phi^{\tK*}}}
 \def\phiF{{\phi^{\tF*}}}
 \def\tJ{\hbox{\tiny \rm J}}
 \def\phiJstar{{\phi^{\tJ*}}}

\def\barX{{\overline {X}}}

\def\trace{\text{\rm trace\,}}

 \newlength{\noteWidth}
\long\def\notes#1{\ifinner
             {\tiny {#1}}
             \else
              \marginpar{\parbox[t]{\noteWidth}{\raggedright\tiny {#1}}}
               \fi}
\def\notes#1{\typeout{#1 !!!}}

\def\spm#1{\notes{spm: #1}}

 \usepackage[usenames]{color}

\def\state{{\sf X}}

\def\U{{\sf U}}

\def\Ebell{{\cal E}_{\hbox{\tiny B}}}

\def\Edir{{\cal E}_{\hbox{\tiny d}}}

\def\fa{\text{\it for all}}

\title{Approximate Dynamic Programming 
\\
using Fluid and Diffusion Approximations
\\
 with Applications to Power Management}

\author{Wei Chen,
Dayu Huang,
Ankur A.\ Kulkarni,
Jayakrishnan Unnikrishnan
\\
Quanyan Zhu,
Prashant Mehta,
 Sean Meyn,
and
Adam Wierman\thanks{A.W. is with Dept.\ of Computer and Mathematical Sciences, California Inst. of Tech.. A.A.K. is with the Systems and Control Engineering group at the Indian Institute of Technology, Bombay, India. J.U. is with the Audiovisual Communications Laboratory, \'{E}cole Polytechnique F\'{e}d\'{e}rale de Lausanne (EPFL), CH-. 1015 Lausanne, Switzerland. 
S.M.\ is with the Dept.\ of ECE,
                University of Florida.
The remaining authors are with   the Coordinated Science Laboratory,
University of Illinois at Urbana-Champaign. Portions of the results presented here were published in abridged form in \cite{chehuakulunnzhumehmeywie09}.}}

\begin{document}

\maketitle

 
\begin{abstract}

Neuro-dynamic programming is a class of powerful techniques for approximating the solution to dynamic programming equations. In their most computationally attractive formulations, these techniques provide the approximate solution only within a prescribed finite-dimensional function class.  Thus, the question that always arises is \textit{how should the function class be chosen}?
The goal of this paper is to propose an approach using the solutions to associated fluid and diffusion approximations.
In order to illustrate this approach, the paper focuses on an application to \textit{dynamic speed scaling} for power management in computer processors.

%
%
%

\end{abstract}

\thispagestyle{empty}




\section{Introduction}
\label{intro}

Stochastic dynamic programming based on controlled Markov chain models have become key tools for evaluating and designing communication, computer, and network applications. These tools have grown in popularity as computing power has increased.  However, even with increasing computing power, it is often impossible to obtain 
exact solutions to dynamic programming problems. This is primarily due to the  so-called ``curse of dimensionality'', which refers to the fact that the complexity of dynamic programming equations often grows exponentially with the size of the underlying state space. 

Recently though, the ``curse of dimensionality'' is slowly dissolving in the face of   approximation techniques such as 
temporal difference learning (TD-learning) and Q-learning \cite{bertsi96a}, which fall under the category of neuro-dynamic programming. These techniques are designed to approximate a solution to a dynamic programming equation within a prescribed finite-dimensional function class. A key determinant of the success of these techniques is the selection of this function class.  Although this question
has been considered in specific contexts in prior work, 
these solutions are often either ``generic'' or highly specific to the application at hand. 
For instance, in \cite{tsiroy97a}, a vector space of polynomial functions has been used   for TD-learning, 
 and a function class generated by a set of Gaussian densities is used in \cite{menmanshi05}. However, determining the appropriate function class for these techniques continues to be more of an art than a science.

\subsection{Main contributions}
The goal of this paper is to illustrate that a useful function class can be designed using solutions to highly idealized approximate models. Specifically, the value functions of the dynamic program obtained under fluid or diffusion approximations of the model can be used as basis functions that define the function class. This can be accomplished by first constructing a fluid or diffusion approximation of the  model, and then solving (or approximating) the corresponding  dynamic programming equation for the simpler system. The value functions of these simpler systems can then be used as some of the basis functions used to generate the function class used in the TD-learning and Q-learning algorithms. 

A significant contribution of this paper is to establish bounds on approximation error by exploiting the similarity between the dynamic programming equation for the MDP model and the corresponding dynamic programming equation for the fluid or diffusion model.  
The bounds are obtained through Taylor series approximations.  A first-order Taylor series approximation is used in a comparison with the fluid model, and a second-order Taylor series is used when we come to the diffusion model approximation.

\subsection{Dynamic speed scaling}

An important tradeoff in modern computer system design is between reducing energy usage and maintaining good performance, in the sense of low delay.   
Dynamic speed scaling addresses this tradeoff by adjusting the processing speed in response to workload. Initially proposed for processor design \cite{BKP07}, 
dynamic speed scaling is now commonly used in many chip designs, e.g. \cite{Inteltech}.
It has recently been applied in other areas such as wireless communication \cite{Kumar04}.  
These techniques have been the focus of a growing body of analytic research \cite{AF06,BPS06,GH01,Wierman09,andrew10}.

For the purposes of this paper, dynamic speed scaling is simply a stochastic control problem -- a single server queue with a controllable service rate -- and the goal is to understand how to control the service rate in order to minimize  a weighted sum of the energy cost and the delay cost.

Analysis of the fluid model provides a good fit to the solution to the average-cost dynamic programming equations,  and subsequent analysis of the diffusion model provides further insight, leading to a two-dimensional basis for application in TD-learning.  
This educated choice for basis functions in TD-learning leads to fast convergence and almost insignificant Bellman error.  A polynomial basis results in much poorer results -- one example is illustrated in \Figure{f:quadraticcostvalue2}.

 Although the paper focuses in large part on the application of TD-learning to the dynamic speed scaling problem, the approach presented in the paper is general:  The use of fluid and diffusion approximations to provide an appropriate basis for  Neuro-dynamic programming is broadly applicable to a wide variety of stochastic control problems.

\subsection{Related work}

Fluid-model approximations  for value functions in network optimization is over fifteen years old \cite{mey97a, mey97b,henmeytad03a,CTCN}, and these ideas have been applied in   approximate dynamic programming   \cite{vea04a,moakumvan06}.   
The same approach is used in \cite{henmeytad03a} to obtain a TD-learning algorithm for variance reduction in simulation for network models.

The preliminary version of this work \cite{chehuakulunnzhumehmeywie09} and the related conference article   \cite{meycheone10} use these techniques for TD-learning,  and motivate the approach through  Taylor series approximations.  In the present work, these arguments are refined to obtain explicit bounds on the Bellman error.

\medbreak

There is a large body of analytic work in the literature studying the dynamic speed scaling problem, beginning with Yao et al.~\cite{YDS95}.  Many focus on models with either a fixed power consumption budget \cite{PvSU05,Bunde06,ZC07} or job completion deadline \cite{PUW04,AF06}.  In the case where the performance metric is the weighted sum of power consumption and delay (as in the current paper), a majority of prior research considers a deterministic, worst-case setting \cite{AF06,BPS06}.  Most closely related to the current paper are \cite{GH01,Wierman09,andrew10}, which consider the MDP described in \eqref{e:powerMDP}.  However, these papers do not consider the fluid or diffusion approximations of the speed scaling model; nor do they discuss the application of TD-learning.

\subsection{Paper organization}

The remainder of the paper is organized as follows.  Some basics of optimal control theory  is reviewed \Section{s.mdp}:   
Markov Decision Process (MDP) models and the associated average-cost optimality equations are reviewed, as well as approaches to \textit{neuro-dynamic programming} for approximating the solution to these equations.   This section also contains formulations of fluid and diffusion models, defined with respect to a general class of MDP models,  along with results explaining why we can expect the solution to one dynamic programming equation might approximate the solution to another.  

 \Section{s.speedscaling} contains application to the power management model, including explicit tight bounds on the Bellman error.  These results are illustrated through simulation experiments in  \Section{s:DSSnum}.  Conclusions are contained in \Section{s.conclusion}. The proofs of all the main results are relegated to the appendix.


\section{Optimal Control}
\label{s.mdp}

We begin with a review of Markov Decision Processes (MDPs).
The appendix contains a summary of  symbols and notation used in the paper.

\subsection{Markov Decision Processes}

\subsubsection*{Dynamic programming equations}

Throughout, we consider the following MDP model.
The state space $\state $ is taken to be $\Re^\ell$, or a subset.   Let $\U\subset\Re^{\ell_u}$ denote the action space, and $\U(x)\in \U$ denote the set of feasible inputs $u$ for $U(t)$ when $X(t)=x$. In addition there is
an i.i.d.\ process $\bfmW$ evolving on $ \Re^w$ that represents
a disturbance process.    For a given initial condition
$X(0)\in\state$, and a sequence $\bfmU$ evolving on $\U$, the
state process $\bfmX$ evolves according to the recursion,
\begin{equation}
X(t+1) = X(t) + f(X(t),U(t), W(t+1)),\qquad t\ge 0.
\label{e:MDP}
\end{equation}
This defines a Markov Decision Process (MDP) with controlled transition law
\[
P_u(x,A) \eqdef \Prob\{ x +  f(x,u, W(1)) \in A\}, \  A\in\bx. 
\]

The controlled transition law can be interpreted as a mapping from functions on $\state$ to functions on the joint state-action space $\state\times\U$:  
For any function $h\colon\state\to\Re$ we denote,
\begin{equation}
P_uh\, (x) =  \Expect[h(X(t+1)) | X(t)=x, U(t)=u].
\label{e:Ptran}
\end{equation}
It is convenient to introduce a \textit{generator} for the model,
\begin{equation}
\generate_uh\, (x)\eqdef \Expect[h(X(t+1)) - h(X(t)) | X(t)=x, U(t)=u].
\label{e:Dgen}
\end{equation}
This is the most convenient bridge between the MDP model and any of its approximations. 

A cost function $c\colon \state\times\U\to \Re_+$ is given.
For a given control sequence $\bfmU$ and initial condition $x=X(0)$, the average cost is given by,
\begin{equation}
\eta^U(x) = \limsup_{n\to\infty}\frac{1}{n}\sum_{t=0}^{n-1}\Expect_x[c(X(t),U(t))],
\label{eta_unr}
\end{equation}
where $\Expect_x[h(X(t))]=\Expect[h(X(t))|X(0)=x]$.  Our goal is to find an optimal control policy with respect to the average cost. The infimum over all $\bfmU$ is denoted $\eta^*$,  which is assumed to be independent of $x$ for this model\footnote{For sufficient conditions see \cite{ber12a,mey97a,mey97b,bor02a,CTCN}.}.


Under typical assumptions \cite{mey97b, mey99a}, an optimal policy
achieving this minimal average cost can be obtained by solving the Average Cost Optimality
Equation (ACOE):
\begin{equation}
\min_{u\in \U(x)} \bigl( c(x,u) + \generate_u h^*\, (x)\bigr) =\eta^*,
\label{e:ACOE}
\end{equation}
where   the generator $ \generate_u$ is defined in \eqref{e:Dgen}.
The ACOE is a fixed point equation in the \textit{relative value
function} $h^*$,  and  the optimal cost for the MDP $\eta^*$ \cite{CTCN}.

For any function $h\colon\state\to\Re$, the $(c, h)$-myopic policy is defined as,
\begin{equation}
\phi^{h}(x)\in\arg\min_{u\in\U(x)}\bigl( c(x,u) + \generate_uh\, (x) \bigr)\, .
\label{e:myopic}
\end{equation}
The optimal policy is a $(c,h^*)$-myopic policy, which is any minimizer,
\begin{equation}
\phi^*(x) \in
\argmin_{u \in \U(x)} \bigl( c(x,u) + \generate_u h^*\, (x)\bigr).
\label{e:ACOEF}
\end{equation}

A related fixed point equation is \textit{Poisson's equation}. It is a degenerate version of the ACOE, in which the control policy is fixed.   Assume a stationary policy $\phi$ is given. Let $\generate= \generate_{\phi}$ denote the resulting generator of the MDP, and $c\colon \state\to \Re$ a cost function. Poisson's equation is defined as,
\begin{equation}
c(x) + \generate h\, (x) =  \eta, \quad x\in \state,
\label{e:fish}
\end{equation}
where  $\eta$ is the average cost defined in \eqref{eta_unr} with policy $\phi$.

\subsubsection*{Approximate dynamic programming}
 
%
%
%

Since solving the dynamic programming equation is complex due to the curse of dimensionality, 
 we use approaches to  approximate a solution to the dynamic programming equation.
  We review standard error criteria next.

The most natural error criterion  is the direct error.  Let $h\colon \state\to\Re$ be an approximation of $h^*$. Define the direct error  $\Edir$ as,
\begin{equation}
\Edir(x):= h^*\, (x)-h\, (x), \quad x\in \state.
\label{e:omega}
\end{equation}
However, the direct error is not
easy to calculate since $h^*$ is not known.

Another common error  criterion is the Bellman error. 
For given $h\colon \state\to\Re$, it is defined as, 
\begin{equation}
\Ebell(x)=\min_{u\in \U(x)}\bigl(c(x,u)+\generate_u h\, (x)\bigr), \quad x\in \state. 
\label{e:Belltheta}
\end{equation}
This is motivated by the associated \textit{perturbed cost function},
\begin{equation}
c^{h}(x,u)=c(x,u)-\Ebell(x)+\eta,
\label{e:ctheta}
\end{equation}
where $\eta\in \Re_+$ is an arbitrary constant. The following proposition can be verified by substituting \eqref{e:ctheta} into \eqref{e:Belltheta}.
\begin{proposition}
\label{t:inve}
The triplet $(c^h,h,\eta)$ satisfies the ACOE,
\[
\min_{u\in\U(x)}\bigl( c^{h}(x,u)+ \generate_u h\, (x) \bigr)  = \eta.
\]
\qed
\end{proposition}

  The solution of the ACOE may be viewed as defining a mapping from one function space to another. Specifically, it maps a cost function $c\colon \state\times\U\to \Re_+$ to a pair $(h^*,\eta^*)$.  Exact computation of the pair ($h^*,\eta^*$) is notoriously difficult, even in simple models when the state space is large.   However, a simpler problem is the inverse mapping problem:  \textit{Given a function $h$ and a constant $\eta$,  find a cost function $c^h$ such that the triple $(c^h,h,\eta)$ satisfies the ACOE \eqref{e:ACOE}}.    The solution of this problem is known as \textit{inverse dynamic programming}.  If the function $c^h$ approximates $c$ then, subject to some technical conditions, the resulting $(c^h,h)$-myopic policy will provide approximately optimal performance.   \Proposition{inve}  indicates that $c^h$ defined in \eqref{e:ctheta} is a  solution to the inverse dynamic programming
problem. 


To evaluate how well $c^h$  approximates $c$, we define the normalized error,
\begin{equation}
\Ecost(x,u)=\frac{|c^h(x,u)-c(x,u)|}{c(x,u)+1}, \quad x\in \state, u\in \U.
\label{e:Bcr}
\end{equation} 
If $\Ecost(x,u)$ is small for all $(x,u)$, then $c^h$ is a good approximation of $c$.


A bound on the Bellman error will imply a bound on the direct error under suitable assumptions.   In most cases the relative value function may be expressed using the \textit{Stochastic Shortest Path} (SSP) representation:
\begin{equation}
 h^*\, (x)=\min_{\bfmU}\Expect_x[\sum_{t=0}^{\tau_{x^\circ}-1}(c(X(t),U(t))-\eta^*)],
 \label{e:acoerepresenta}
\end{equation}
where $\tau_{x^\circ}=\min(t\geq1: X(t)=x^\circ)$ is the first return time to a state $x^\circ$.
Conditions under which the SSP representation holds are given in  \cite{mey99a, mey97b, ber91,ber12a,CTCN}.


 The following result establishes bounds on the direct error.


\begin{proposition}
\label{t:errorboundh}
Suppose that $h^*$ and $h$ each admit SSP representations.  For the latter, this means that
\begin{equation}
 h\, (x)=\min_{\bfmU}\Expect_x[\sum_{t=0}^{\tau_{x^\circ}-1}(c^h(X(t),U(t))-\eta)],\nonumber
\end{equation}
where $c^h$ and $\eta$ are defined in \eqref{e:ctheta}, and the corresponding policy is the $(c^h,h)$-myopic policy.
Then, the following upper and lower bounds hold for the direct error,
\begin{equation}
\begin{aligned}
\Edir(x) \geq &\Expect_x^{\phi^*}[\sum_{t=0}^{\tau_{x^\circ}-1}(\Ebell(X(t))-\eta^*)],\\
\Edir(x) \leq& \Expect_x^{\phi^{h}}[\sum_{t=0}^{\tau_{x^\circ}-1}(\Ebell(X(t))-\eta^*)].\nonumber
\end{aligned}
\end{equation}
where $\Expect^\phi[h(X(t))]$ denotes the expectation taken over the controlled Markov chain under policy $\phi$. 
\end{proposition}
The bounds for the direct error in \Proposition{errorboundh} are derived by using the Bellman error and the SSP representation.  The details of the proof are given in \Section{s:Error bounds} of the appendix.

\subsection{Neuro-dynamic programming}
\label{s:pappro}

Neuro-dynamic programming concerns approximation of dynamic programming equations through either simulation or through observations of input-output behavior in a physical system.  The latter is called reinforcement learning, of which TD-learning is one example. 
The focus of this paper is on TD-learning, although similar ideas are also applicable for other approaches such as Q-learning.

Approximation of the relative value function will be obtained with respect to a  parameterized function class: $\clF = \{h^{\theta} : \theta\in \Re^d\}$.
In TD-learning a fixed \textit{stationary} policy is considered (possibly randomized),  
and the goal is to find the parameter $\theta^*$ so that $h^{\theta^*}$ best approximates the  solution $h$ to  Poisson's equation \eqref{e:fish}.   In standard versions of the algorithm, the direct error is considered.

TD-learning algorithms are based on the assumption that the given policy is stabilizing, in the sense that the  Markov chain has unique   invariant probability distribution $\pi$. 
The mean-square direct error is minimized,  
\begin{equation}
\Expect_\pi[\bigl(h(X(0)) - h^{\theta}(X(0))\bigr)^2]  =\int \bigl(h\, (x) - h^{\theta}(x)\bigr)^2\, \pi(dx)\nonumber
\end{equation} 
In the original TD-learning algorithm, The optimal parameter is obtained through a stochastic approximation algorithm based on steepest descent.   The LSTD (least-squares TD) algorithm is a Newton-Raphson stochastic approximation algorithm. 

TD-learning was first introduced in \cite{brabar96}. The  LSTD-learning for the average cost problem is described in \cite{kontsi03a,CTCN}.


In this paper we restrict to a linear function class.  It is assumed that there are $d$ basis functions,  $\{\psi_i \colon\state\to\Re, 1\le i\le d\}$ so that   $\clF=\{h^{\theta}:=\sum_{i=1}^d{\theta_i}\psi_i\}$.  We also write $h^\theta=\theta^\transpose \psi$.  In this special case, 
the optimal parameter $\theta^*$ is the solution to a least-squares problem, and the LSTD algorithm is frequently much more reliable than other approaches, in the sense that variance is significantly reduced.

The TD-learning algorithm is used to compute an approximation of the relative value function for a specific policy.  To estimate the relative value function with the optimal policy, the TD-learning algorithm is combined with policy improvement.

The policy iteration algorithm (PIA) is a method to construct an optimal policy through the following steps.  The algorithm is initialized with a policy $\phi^0$ and then the following operations are performed in  the $k$th stage of the algorithm:
\begin{romannum}
\item Given the policy $\phi^k$, find the solution $h^k$ to
    Poisson's equation $\generate_{\phi^k} h^k  + c_k  = \eta_k$,   where
    $c_k(x) = c(x,\phi^k(x))$,   and $\eta_k$ is the average
    cost.
\item  Update the policy via
\[
\phi^{k+1}(x) \in \argmin_{u\in\U(x)}\{ c(x,u) + \generate_u h^k\, (x)\}.
\]
\end{romannum}

In order to combine TD-learning with PIA, the TDPIA algorithm considered replaces the first step with an application of the LSTD algorithm, resulting in an approximation $h^k_{\td}$ to the function   $h^k$.   The policy in (ii) is then taken to be   $\phi^{k+1}(x)\in \argmin_{u\in\U(x)} \{ c(x,u)+\generate_u h^{k}_{\td}(x) \}$.

\subsection{Approximation architectures}
\label{s.approach}

We now introduce approximate models and the corresponding approximations to the ACOE. 

\subsubsection*{Fluid model}

The fluid model associated with the MDP model given in \eqref{e:MDP} is defined by the ordinary differential equation,
\begin{equation}
\ddt x(t) = \barf(x(t),u(t)),\qquad x(0)\in\state,
\label{e:fm}
\end{equation}
where $\barf(x,u) \eqdef
\Expect[f(x,u, W(1))]$.
The fluid model has state $\bfmx$ that evolves on $\state$,  and input  $\bfmu$ that evolves on $\U$.  The existence of  solutions to \eqref{e:fm} will be assumed throughout the paper. 

The fluid value function considered in prior work on network models surveyed in the introduction is defined to be the infimum over all policies of the total cost,  
\begin{equation}
J^*(x) = \inf_{\bfmu} \int_0^\infty c(x(t),u(t))  \, dt,\qquad x(0)=x\in\state
\label{e:Jstar}
\end{equation}
However, in the general setting considered here, there is no reason to expect that $J^*$ is finite-valued.

In this paper we consider instead the associated perturbed HJB equation:  For given $\eta>0$,  we assume we have a solution to the differential equation,
\begin{equation}
\min_{u\in\U(x)}\bigr(c(x,u)  +\generate_u^FJ^*(x)\bigr) = \eta.
\label{e:hfirst1}
\end{equation}
where $\generate_u^F$ is the generator for the fluid model defined as:
\begin{equation}
\generateF_uh\, (x)
=
\ddt h(x(t))\Big|_{t=0, u(0)=u, x(0)=x} = \nabla h\, (x) \cdot \barf(x,u).   
\label{e:GenFa}
\end{equation}
Given a solution to \eqref{e:hfirst1},  we denote the corresponding policy for the fluid model, 
\begin{equation}
\phiF(x) \in
\argmin_{u\in\U(x)} \bigl( c(x,u)+\generate_u^FJ^*(x)\bigr).
\label{e:TCOEF}
\end{equation}
 In the special case that the total cost \eqref{e:Jstar} is finite-valued,  and some regularity conditions hold,  this value function will solve \eqref{e:hfirst1} with $\eta=0$,
 and $\phiF$ will be an optimal policy with respect to total-cost \cite{ber05a}.

 The HJB equation \eqref{e:hfirst1} can be interpreted as an optimality equation for an optimal stopping problem for the fluid model.  Denote the first stopping time   $T_{\eta}=\min\{t: c(x(t),u(t) )\leq \eta \}$, and let $K^*_\eta$ denote the minimal total \textit{relative cost},
\begin{equation}
K_\eta^*(x) = \inf_{\bfmu} \int_0^{T_\eta} \bigl(c(x(t),u(t) )-\eta\bigr) \, dt,
\label{e:Ketaw}
\end{equation}
where the infimum is over all policies for the fluid model. Under reasonable conditions this function will satisfy the dynamic programming equation,
\begin{equation}
\min_{u \in \U(x)}\bigl( c(x,u) -\eta  + \generateF_u K_\eta^*\, (x) \bigr) = 0.\nonumber
\end{equation}
A typical necessary condition for finiteness of the value function is that  the initial condition $x$ satisfy $\min_u c(x,u)>\eta$.

Let us now investigate the relationship between the MDP model and its fluid model approximation.  It is assumed that $J^*$ is a smooth solution to \eqref{e:hfirst1}, so that the following first-order Taylor series expansion is justified:  Given $X(0)=x$,  $ U(0)=u  $,
\begin{equation*}
\begin{aligned}
 \generate_uJ^*\, (x) &\approx \Expect 
\bigl[\nabla J^*(X(0)) (X(1) - X(0))  ]
\\   
 & =\nabla J^*\, (x) \cdot \barf(x,u)\\   
 & = \generateF_uJ^*
\end{aligned}
\end{equation*}
A more quantitative approximation is obtained when $J^*$ is twice continuously differentiable ($C^2$).

If  $J^*$ is of class $C^2$ then  lower and upper bounds on the Bellman error $\Ebell$ in the following proposition, which measure the quality of the approximation of $J^*$ to $h^*$. The proof of   \Proposition{clEbdds} is contained in the appendix.

\begin{proposition}
\label{t:clEbdds}
Suppose that   $J^*$ is a $C^2$ solution to \eqref{e:hfirst1}.
Then, the Bellman error admits the following  bounds for each $x$, 
\begin{romannum}

\item
With
$\phiJstar(x)\in\arg\min_{u\in\U(x)}\bigl( c(x,u) + \generate_uJ^*\, (x) \bigr)$   the $(c,J^*)$-myopic policy, 
and   $\Delta_X =   f(x,\phiJstar(x),W(1))$, 
\[
 \Ebell(x) \geq\half  \Expect [ \Delta_X^\transpose \nabla^2J^*\, (\barX_l) \Delta_X ]
+\eta, 
\]
where the random variable $\barX_l$ takes value between $x$ and $x+\Delta_X$.

 \smallbreak

\item 
 With  $\phiF(x)$ is the policy given in \eqref{e:TCOEF},
and $\Delta_X=f(x,\phiF(x),W(1))^\transpose$,
\[ 
 \Ebell(x) \leq \half  \Expect [ \Delta_X^\transpose   \nabla^2J^*\, (\barX_u) \Delta_X]+\eta,
 \]
where the random variable $\barX_u$  takes value between $x$ and $x+f(x,\phiF(x), W(1))$.

\end{romannum}

\end{proposition}

\subsubsection*{Diffusion model}

The diffusion model is a refinement of the fluid model to take into account volatility.  It is
motivated similarly, using  a \textit{second-order} Taylor series expansion.

To bring in randomness to the fluid model, it is useful to first introduce a  fluid model in discrete-time.   For any $t\ge 0$, the random variable denoted $\Delta(t+1)  = f(X(t),U(t), W(t+1)) - \barf(X(t),U(t))$ has zero mean.   The evolution of $\bfmX$ can be expressed as a discrete time nonlinear system, plus  ``white noise'',
\begin{equation}
X(t+1) = X(t) + \barf(X(t),U(t)) +\Delta(t+1),\qquad t\ge 0.
\label{e:MDPbarfDelta}
\end{equation}
The fluid model in discrete time is obtained by ignoring the noise,
\begin{equation}
x(t+1) = x(t) + \barf(x(t),u(t)).
\label{e:MDPbarf}
\end{equation}
This is the discrete time counterpart of \eqref{e:fm}.  Observe that it can be justified exactly as in the continuous time analysis:  If $h$ is a smooth function of $x$, then we may approximate the generator using a first order-Taylor series approximation as follows:
\begin{equation}
\begin{aligned}
\generate_u h\, (x)  &\eqdef \Expect[h(X(t+1)) - h(X(t)) \mid X(t)=x, U(t)=u]
\\
& \approx
  h(x + \barf(x,u))  - h\, (x)
\\
&
 +
  \Expect   [\nabla h(x + \barf(x,u)) \cdot \Delta(1) \mid X(0)=x, U(0)=u  ]
  \\
 &=
  h(x + \barf(x,u))  - h\, (x)\nonumber
\end{aligned}
\end{equation}
The right hand side is the  generator applied to $h$, for the \textit{discrete-time} fluid model
\eqref{e:MDPbarf}.

Consideration of the model
\eqref{e:MDPbarf} may give better approximation for value functions in some cases, but we lose the simplicity of differential equations that characterize value functions in the continuous time model \eqref{e:fm}.

The representation \eqref{e:MDPbarfDelta}  also motivates the diffusion model.   Denote the conditional covariance matrix   by
\[
\Sigma_f(x,u) = \Expect[\Delta(t+1)\Delta(t+1)^\transpose \mid X(t)=x,\, U(t)=u]\, ,
\]
and let $b$ denote an $\ell\times\ell$ ``square-root'', $b(x,u)b(x,u)^\transpose = \Sigma_f(x,u)$ for each $x, u$.
The diffusion model is of the form,
\begin{equation}
dX(t) = \barf(X(t),U(t))\, dt + b(X(t),U(t)) \, dN(t)
\label{e:diffModel}
\end{equation}
where the process $\bfmN$ is a standard Brownian motion on $\Re^\ell$.

To justify its form we consider a second order Taylor series approximation.
If $h\colon\state\to\Re$ is a $C^2$ function, and at time $t$ we have  $X(t)=x$, $U(t)=u$,  then the standard second order Taylor series about  $x^+ =  x+\barf(x,u)$ gives,
\begin{equation}
\begin{aligned}
&h(X(t+1)) - h(X(t)) \\
\approx &\nabla h\,(x^+) \cdot  \barf(x,u)  + \half  \Delta(t+1)^\transpose  \nabla^2 h\, (x^+)  \Delta(t+1).\nonumber
\end{aligned}
\end{equation}
Suppose we can justify a further approximation, obtained by replacing $x^+$ with $x$ in this equation.  Then, on
taking expectations of each side, conditioned on $X(t)=x$, $U(t)=u$, we  approximate the generator for $\bfmX$ by the   second-order ordinary differential operator,
\begin{equation}
 \generateD_u h\, (x)  \eqdef \nabla h\, (x) \cdot \barf(x,u)+ \half    \trace \Bigl(\Sigma_f(x,u) \nabla^2 h\, (x) \Bigr)\, .
\label{e:diffusionGen}
\end{equation}
This is precisely the differential generator for \eqref{e:diffModel}.

The minimal average cost $\eta^*$ is defined as in the MDP model  \eqref{eta_unr}, but as a continuous-time average.
The ACOE has the precise form as in the MDP, only the definition of the generator is changed:
\begin{equation}
\min_{u \in \U(x)}\bigl( c(x,u) + \generateD_u h^*\, (x) \bigr) = \eta^*
\label{e:ACOED}
\end{equation}
and $h^*$ is again called the \textit{relative value function}.

The solution to the ACOE for the diffusion model is often a good approximation for the MDP model.  Moreover, as in the case of fluid models,  the continuous time model is more  tractable because tools from calculus can be used for computation or approximation.  This is illustrated in the power management model in the next section.

\section{Power management model}
\label{s.speedscaling}

The
dynamic speed scaling
problem surveyed in the introduction
is now addressed through the techniques described in the previous section.
To begin, we construct an MDP that  is described as a single server queue with a controllable service rate that determines  power consumption.

\subsection{The MDP model}
\label{s.speedscalingMDP}

For each $t=0,1,2,\dots$,   let $A(t)$ denote the job
arrivals in this time slot,  $X(t)$ the number of jobs in the queue awaiting
service,   and  $U(t)$ the rate of service. The MDP model is a controlled random walk:
\begin{equation}
X(t+1)=X(t) -   U(t) + A(t+1),\qquad t\ge 0.
\label{e:powerMDP}
\end{equation}

The following assumptions on the arrival   process are imposed throughout the paper.
\def\As#1{{\textbf{A#1}}}
\begin{romannum}
\item[\As{1}]
The arrival process $\bfmA$ is i.i.d.,  its marginal distribution is supported on $\Re_+$ with finite mean $\alpha$.  Moreover, 
zero is in the support of its distribution:
\[
\Prob\{ A(1) = 0\}>0, 
\] 
and there exists a constant $N>0$ such that $\Prob\{ A(1)<N\}=1$. 
\end{romannum}
The boundedness assumption is not critical -- a $p$th moment for $p>2$ would be sufficient.   The assumption that $A(t)$ may be zero is imposed to facilitate a proof that the controlled Markov model is ``$x^\circ$-irreducible'', with $x^\circ=0$  (see \cite{CTCN}).

The cost function is chosen to balance cost of delay with power consumption:
\begin{equation}
c(x,u) = x + \nu \clP(u),\nonumber
\label{e:cP}
\end{equation}
where $\clP$ denotes the power consumption as a function of the service rate $u$, and $\nu>0$.  This form of cost function is common in the literature, e.g., \cite{GH01,Wierman09, andrew10}.   

The remaining piece of the model is to define the form of $\clP$.  Two forms of $\clP$ are considered, based on two different applications: processor design and wireless communication.

For processor design applications, $\clP$ is typically assumed to be a polynomial. In earlier work in the literature, $\clP$ has been taken to be a cubic. The reasoning is that the dynamic power of CMOS is proportional to
$V^2 f$, where $V$ is the supply voltage and $f$ is the clock
frequency~\cite{KM08}.  Operating at a higher frequency
requires dynamic voltage scaling (DVS) to a higher voltage,
nominally with $V \propto f$, yielding a cubic relationship.
 However, recent work, e.g.\ \cite{Wierman09}, has found that
the dynamic power usage of real chips is well modeled by a
polynomial closer to quadratic.  When considering polynomial cost, 
in this paper we take a single term,
\begin{equation}
 \clP(u) = u^\varrho
 \label{eq:power-scaling}
\end{equation}
where $\varrho > 1$, and  we focus primarily on the particular case of $\varrho=2$.

For wireless communication applications, the form of $\clP(u)$ differs significantly for different scenarios. An additive white Gaussian noise model  \cite{Kumar04} gives,
\begin{equation}
 \clP(u) = e^{\kappa u}
 \label{eq:power-exponential-scaling}
\end{equation} for some
$\kappa>0$.

Considered next are   fluid and diffusion models to approximate the solution to the ACOE in this application.

\subsection{Convex and continuous approximations}

Here we apply the fluid and diffusion approximations introduced in \Section{s.approach} to the power management model, and derive explicit error bounds for the approximations to the ACOE.  

\subsubsection*{The fluid model}


Specializing the fluid model given in \eqref{e:fm} to the case of dynamic speed scaling \eqref{e:powerMDP} gives the following continuous-time deterministic model:
\begin{equation}
\ddt x(t) = -  u(t) + \alpha,\nonumber
\end{equation}
where $\alpha $ is the expectation of $A(t)$,  and the processing speed $u(t)$ and queue length $x(t)$ are both non-negative.

The total cost $J^*(x)$ defined in \eqref{e:Jstar} is not finite for the cost function 
\eqref{e:cP} when $\clP$ is
defined in \eqref{eq:power-scaling} or \eqref{eq:power-exponential-scaling}. 
Consider the alternate value function   defined in \eqref{e:Ketaw},
with $\eta=0$ and with the modified stopping time,
\[
T_0 \eqdef \inf_t\{x(t)=0 \}
\]
This value function is denoted, for $x(0)=x\in \Re_+$, by
\begin{equation}
K^*(x) = \inf _{\bfmu}\int_0^{T_0} c(x(t),u(t) ) \, dt, 
\label{e:Keta}
\end{equation}
where the infimum is over all input processes for the fluid model. 
The function $K^*$ solves the HJB equation \eqref{e:hfirst1} with $\eta=0$, and is finite-valued. 

Fluid model with general polynomial cost  are discussed in detail in \Section{s:fluidandprop}.  Here we review results in the simple case of quadratic cost with $\nu = \half$,
\begin{equation}
c(x,u) = x + \half  u^2\, .
\label{e:QuadA}
\end{equation}
In this case we obtain by elementary calculus,
\begin{equation}
K^*(x) = \alpha x + \third [(2x + \alpha^2)^{3/2} - \alpha^3].
\label{e:KstarQuadA}
\end{equation}

Rather than modify the optimization criterion, we can approximate the solution to
\eqref{e:hfirst1}  by modifying the cost function  $c(x,u)$  
so that it vanishes at the equilibrium $(x=0, u=\alpha)$. 
A simple modification of \eqref{e:QuadA} is the following,
\[
c(x,u) = x + \half  ( [u- \alpha]_+)^2\, ,
\]
where $[\varble]_+ = \max(0,\varble)$.  The fluid value function  $J^*$ defined in \eqref{e:Jstar} is similar to $K^*$ given in \eqref{e:KstarQuadA},
\begin{equation}
J^*(x)  = \third (2 x)^{3/2}
\label{e:quadJ}
\end{equation}
and in this case the optimal fluid policy takes the simple form
   $\phiF(x)=\sqrt{2x}+\alpha$,

To measure the quality of these approximation   to the relative value function $h^*$, we estimate bounds for the Bellman error and direct error.  
In  \Proposition{clEpos}, we first compute the bounds for the Bellman error. They are derived based on \Proposition{clEbdds}. The bounds for the direct error are derived based on \Proposition{errorboundh}, which assumes the SSP representation holds for $h^*$.  This will be assumed throughout our analysis:

\As{2} 
The relative value function $h^*$   admits the SSP representation \eqref{e:acoerepresenta}, with $x^\circ=0$.   Moreover, under the optimal policy, any bounded set is \textit{small}:  For each $N\ge 1$ there exists $T<\infty$ and $\epsy>0$  such that whenever $ x\le N$ and $t\ge T$,
\begin{equation}
\Prob\{ X(t) = 0 \mid X(0)=x\}\ge \epsy.
\label{e:small}
\end{equation}

Analytic techniques to verify this assumption are contained in  \cite{wei2013} (see also 
 \cite{chemey99a} for verification of the ``small set'' condition \eqref{e:small} for countable state-space models).

Under these assumptions we can obtain bounds on the Bellman error.  We present explicit bounds only for the case of quadratic cost.
\begin{proposition}
\label{t:clEpos}
Consider the speed scaling model with cost function \eqref{e:QuadA},
and with $K^*$ the fluid value function given in \eqref{e:KstarQuadA}.
The following hold for the Bellman error $\Ebell$ defined in \eqref{e:Belltheta}
and the   direct error $\Edir$ defined in \eqref{e:omega}: 
\begin{romannum}
\item
Under Assumption~\As{1}, the Bellman error
 is non-negative, and grows at rate $\sqrt{x}$, with
 \[
\lim_{x\rightarrow \infty} \frac{\Ebell(x)}{\sqrt{x}}  = \frac{1}{\sqrt{2}}.
\] 

\item
If in addition Assumption \As{2} holds,
then the direct error $\Edir$     satisfies, 
\[
\underline{\Edir}(x)\leq \Edir(x)\leq \overline{\Edir}(x),
\]
where  the upper bound grows at most linearly, $\overline{\Edir}(x)=\mathcal{O}(x)$,  and the lower bound is independent of $x$; that is,  $\underline{\Edir}(x)=\mathcal{O}(1)$.
\end{romannum}
In particular, under Assumptions \As{1} and  \As{2}, 
\[
\lim_{x\to\infty} \frac{ \Ebell(x)}{c(x,0)} = 
\lim_{x\to\infty} \frac{ \Edir(x)}{K^*(x)} = 0. 
\]

\end{proposition}

The bounds on the Bellman error in (i) imply a bound on the    normalized error defined in \eqref{e:Bcr}:
\begin{equation}
\begin{aligned}
\lim_{x\to\infty}\Ecost(x,u)&=\lim_{x\to\infty}\frac{|c(x,u)-c^{K^*}(x,u)|}{c(x,u)+1}\\
&
\
=
\lim_{x\to\infty}\frac{|\eta-\Ebell(x)|}{c(x,u)+1}=0, \quad \fa \, u\in \U.
\end{aligned}
\label{e:Ecostlim}
\end{equation}
 This implies that the inverse dynamic programming solution $c^{K^*}$ gives a good approximation of the original cost for large $x$.
 The bounds on the direct error in (ii) imply that the fluid value function is a good approximation for large $x$, 
\[
\lim_{x\to\infty}\frac{K^*(x)}{h^*\, (x)}=1.
\]

\subsubsection*{The diffusion model}

Following \eqref{e:diffusionGen}, the generator for the  diffusion model is expressed, for any smooth function $h$ by,
\begin{equation}
\generate_u^D h= (-u+\alpha)  \nabla h    +\half   \sigma_A^2 \nabla^2 h ,
\label{e:generatordiff}
\end{equation}
where $\sigma_A^2=\Expect[(A(1)-\alpha)^2]$.

Provided $h^*$ is monotone, so that $\nabla h^*\, (x) \geq 0$ for all $x$,  the minimizer in  \eqref{e:ACOED} can be expressed
\begin{equation}
\phi^*(x)=\nabla h^*\, (x)\, .
\label{e:control}
\end{equation}
Substituting \eqref{e:control} into \eqref{e:ACOED} gives the nonlinear equation,
\begin{equation}
x-\half  (\nabla h^*\, (x))^2+\alpha \nabla h^*\, (x)+\half   \sigma_A^2 \nabla^2 h^*\, (x)-\eta^*=0.
\label{e:fixedpointdiffusion}
\end{equation}
We do not have a closed form expression, but we will show that an approximation is obtained as a perturbation of
$K_\eta^*$ defined in \eqref{e:Ketaw}, with $\eta=\eta^*$.  The constant $\eta^*$ is not known, but the structure revealed here will allow us to obtain the desired basis for TD-learning.

For any function $h\colon\Re_+\to\Re$, denote the Bellman error for the diffusion model by,
\begin{equation}
\Ebell^\tD(x)\eqdef \min_{u\in \U(x)}\bigl(c(x,u)+\generateD_u h\, (x)\bigr), \quad x\in \state.
\label{e:Bellh}
\end{equation}
The value function $K^*$  defined in \eqref{e:KstarQuadA} is an approximation of $h^*$ in the sense that the Bellman error is bounded:
\begin{proposition}
\label{t:DiffBellman0}
The function $K^*$ defined in \eqref{e:KstarQuadA} is convex and increasing.  Moreover,  with $h=K^*$,
the Bellman error for the diffusion has the explicit form,
$\Ebell^\tD=\half  \sigma_A^2 (2x+\alpha^2)^{-\half }$.
\end{proposition}

A tighter approximation can be obtained with the `relative value function' for the fluid model,
$K_\eta^*$.  This satisfies a nonlinear equation similar to \eqref{e:fixedpointdiffusion},
\[
x-\half  (\nabla K_\eta^*\,(x))^2+\alpha \nabla K_\eta^*\,(x)-\eta =0,\quad x> \eta.
\]
Elementary calculations show that for any constant $q>0$ we have the approximation,
\[
\nabla K_\eta^*\,(x)  \approx \nabla K^*\,(x)  -   \eta (2x +q^2)^{-\half } +O((1+x)^{-3/2}),
\]
where $K^*$ is given in \eqref{e:KstarQuadA}.    We then take the right hand side for granted in the approximation
$\nabla h\,(x)  \eqdef \nabla K^*\,(x)  -   \eta (2x +q^2)^{-\half }  $,  and on integrating this gives
\begin{equation}
h\, (x) = K^*(x) -  \eta  \sqrt{2x +q^2}  +  \eta   q.
\label{e:diffusionbasis}
\end{equation}
The constant is chosen so that $h(0)=0$.

We note that this is a reflected diffusion, so that   the generator is subject to the boundary condition  $\nabla h\,(0) =0$  \cite[Theorem 8.7.1]{CTCN}. 
The boundary condition is satisfied for \eqref{e:diffusionbasis} on taking  $q\eqdef  \eta/\alpha$.

The resulting approximation works well for the diffusion model.
The proof of \Proposition{DiffBellman} follows from calculus computations contained in the Appendix.
\begin{proposition}
\label{t:DiffBellman}
Assume that $\nabla h(0) \ge 0$. For any $\eta>0$, $q>0$, the function $h$ defined in \eqref{e:diffusionbasis} is convex and increasing.  Moreover,
the Bellman error $\Ebell^\tD$  is bounded over $x\in\Re_+$, and the following bound holds for large $x$:
\[
|\Ebell^\tD(x) -  \eta | = O((1+x)^{-\half}).
\]
\qed
\end{proposition}

The function \eqref{e:diffusionbasis} also serves as an approximation for the MDP model.  The proof is omitted since it is similar to the proof of \Proposition{clEbdds}.

\begin{proposition}
\label{t:clEbddsDiff}
The function $h$ defined in \eqref{e:diffusionbasis} is an approximate solution to the ACOE for the MDP model, in the sense that  the Bellman error   has growth of order $\sqrt{x}$,  
\[
\lim_{x\rightarrow \infty} \frac{\Ebell(x)}{\sqrt{x}}   = \frac{1}{\sqrt{2}}
\] 
\qed
\end{proposition}

The bounds obtained in this section quantify the accuracy of the fluid and diffusion model approximations. We next apply this insight to design the function classes required in TD-learning algorithms.

\section{Experimental results}
\label{s:DSSnum}

The results of the previous section are now illustrated with results from numerical experiments.  We restrict to the quadratic cost function
given in  \eqref{e:QuadA} due to limited space.

In application to TD-learning, based on a linear function class as discussed in \Section{s:pappro}, the following two dimensional basis follows from the analysis of the fluid and diffusion models,
\begin{equation}
\psi_1(x) = K^*(x),\quad \psi_2(x) =  q-\sqrt{2x +q^2}     ,\qquad x\ge 0,
\label{e:PolyBasis}
\end{equation}
where the value function  $K^*$ is given in \eqref{e:KstarQuadA}.
The basis function $\psi_1$ is motivated by the error bound in \Proposition{clEpos},  
and $\psi_2$ is motivated by \Proposition{clEbddsDiff}.  
The parameter $\eta\ge 0$ is fixed, and then we take  $q=\eta/\alpha$


The details of the simulation model are as follows: The arrival process $\bfmA$ is i.i.d.\ on $\Re_+$, satisfying the assumptions imposed in
 \Section{s.speedscalingMDP}, although the boundedness assumption in \As{1} was relaxed for simplicity of modeling:
   In the first set of experiments,
 the marginal distribution is a
 scaled geometric random variable, of the form
\begin{equation}
A_0(t) =\Delta_{A_0} G(t),\qquad t\ge1,
\label{e:A1}
\end{equation}
where $\bfmG$ is geometrically distributed on $\{0,1,\dots\}$ with parameter $ p_{A_0}=0.96$, and $\Delta_{A_0}$ is chosen so that the mean $\alpha_0$ is equal to unity:
\begin{equation}
 1= \alpha_0 = \Delta_{A_0}\frac{p_{A_0}}{1-p_{A_0}}\ \  \hbox{\it and} \ \  \Delta_{A_0}=1/24. 
\label{e:pA1}
\end{equation}
The variance of $A_0$ is given by,
\[
\sigma^2_{A_0} =
\frac{p_{A_0}}{(1-p_{A_0})^2}\Delta_{A_0}^2=1.
\]
 In  \Section{s:Arho}  experiments are described in which the variance of the arrival process was taken as a parameter, to investigate the impact of variability.

\subsection{Value iteration}
\label{s.VIA}

We begin   by computing  the actual solution to the average
cost optimality equation using value iteration. This provides a
reference for evaluating the proposed approach for TD-learning.

The value iteration (VIA) solves the fixed point equation \eqref{e:ACOE} based on the successive approximation method -- see \cite{whi63} for the origins,  and  \cite{chemey99a,CTCN} for more recent relevant results.

The algorithm is initialized with a function $V_0$, and the iteration is given by,
\begin{equation}
V_{n+1}(x)=\min_{u\in\U(x)}\bigl(c(x,u)+P_uV_n(x)\bigr), \quad n\geq0 .
\label{e:viaupdate}
\end{equation}
For each $n\geq 1$, the function $V_n$ can be expressed as
\begin{equation}
V_n(x)=\min_{\bfmU}\Expect_{x}[\sum_0^{n-1} c(X(t),U(t)) +V_0(X(n))].
\label{e:val_lim}
\end{equation}

\begin{figure}[t]
\Ebox{1.0}{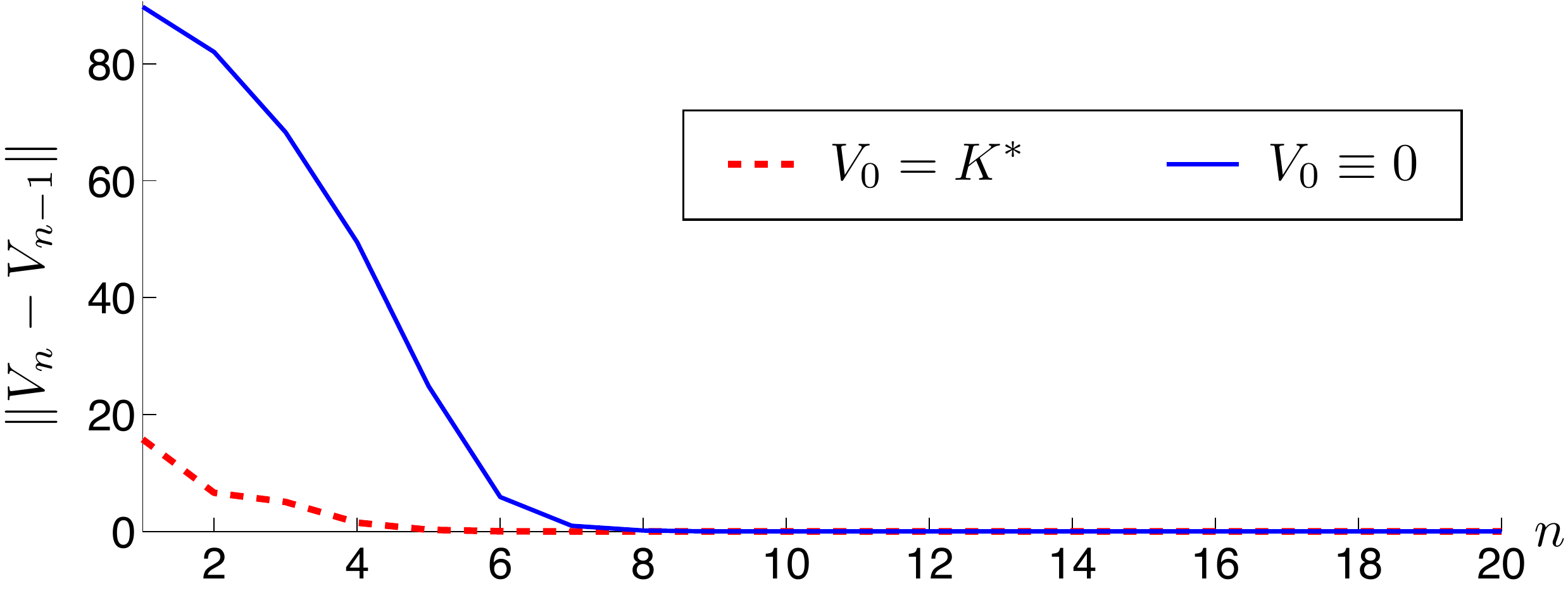}
\caption{The convergence of value iteration for the quadratic cost function \eqref{e:QuadA}.}
\label{f:Error1}
\end{figure}

The approximate solution to the ACOE at stage $n$ is taken to be the normalized value function $h_n(x)=V_n(x)-V_n(0)$,  $x\in\state$, where $V_n$ is the $n$th value function defined in \eqref{e:val_lim}.  The convergence of $\{h_n\}$ to $h^*$ is illustrated in \Figure{f:Error1}. The error  $\| h_{n+1} - h_n\|$ converges to zero \textit{much faster} when the algorithm is initialized using the fluid value function of \eqref{e:KstarQuadA}. 

\begin{figure}[ht]
\Ebox{1.0}{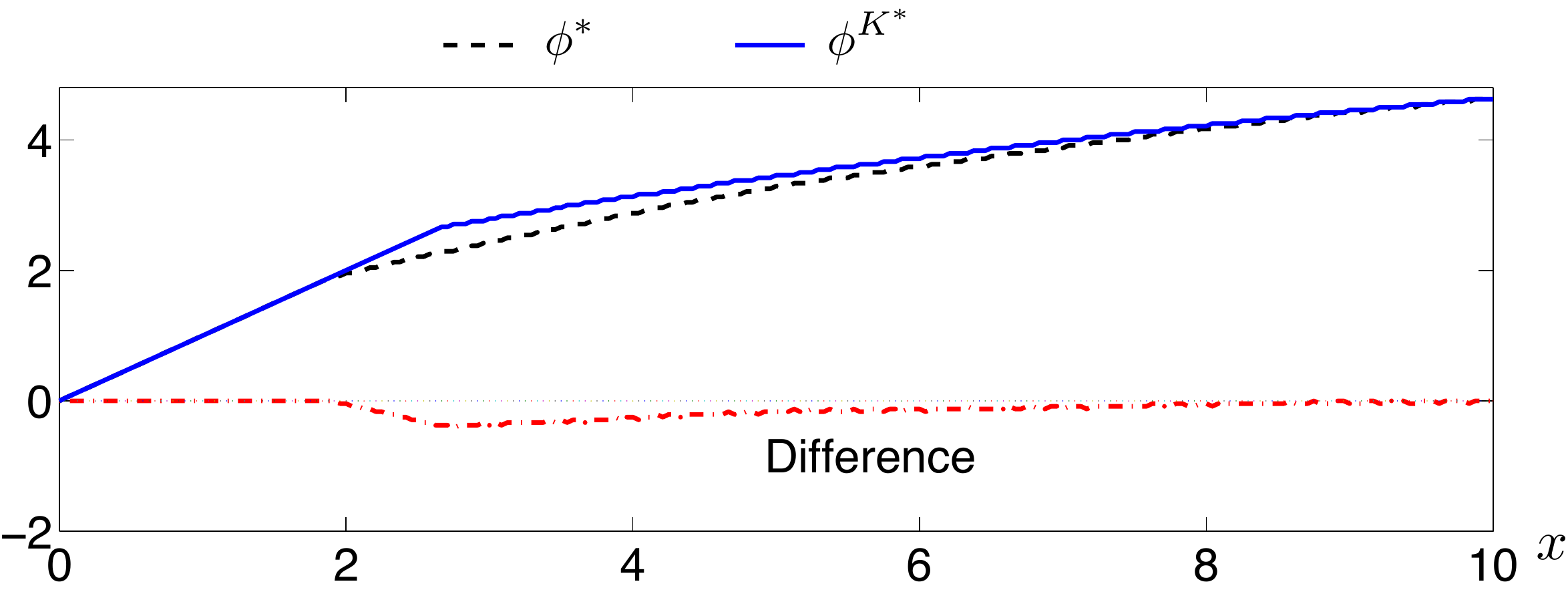}
\caption{The optimal policy $\phi^*$ compared to the $(c,K^*)$-myopic policy, $\phi^{K^*}$, for
the quadratic cost function $c(x,u) = x+\half  u^2$.}
\label{f:Policies1}
\end{figure}

Shown in \Figure{f:Policies1} is a comparison of the optimal  policy $\phi^*$, computed numerically using value iteration, and the   $(c,K^*)$-myopic policy, $\phi^{K^*}$. 

\subsection{TD-learning with policy improvement}
\label{s.TDexp}

The first set of experiments illustrates convergence of the TD-learning algorithm with policy improvement introduced in \Section{s:pappro}.

Recall from \Proposition{clEbddsDiff}  
that a linear combination of the basis functions in \eqref{e:PolyBasis} provides a tight approximation to the ACOE for the MDP model. 
In the numerical results surveyed here it was found that the average cost is approximated by $\eta^*\approx 2$, and recall that we take  $\alpha=1$ in all experiments.    
Hence $\eta=2$ and $q=\eta/\alpha=2$ were chosen in the basis function $\psi_2$ given in  \eqref{e:PolyBasis}.

The initial policy was taken to be $\phi^0(x)= \min(x,1)$,  $x\ge 0$, and the initial condition for TD learning was taken to be $\theta(0) = (0,0)^\transpose$.

\begin{figure}[ht]
\Ebox{1.0}{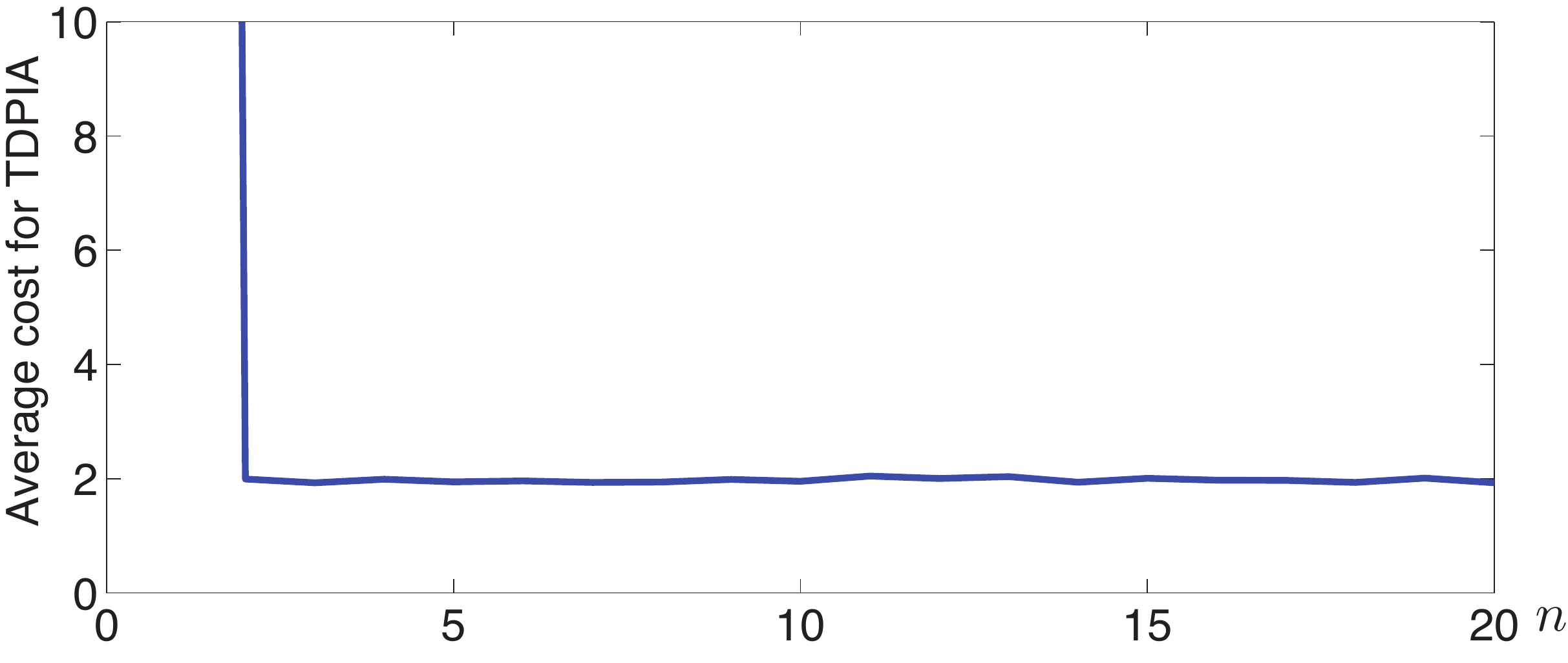}
\caption{Simulation result for TDPIA  with the quadratic cost function \eqref{e:QuadA},
and   basis given in \eqref{e:PolyBasis}.}
\label{fig:TDPIA1}
\end{figure}

 \Figure{fig:TDPIA1} shows the estimated average cost in each of the twenty iterations of the algorithm.   The algorithm results in a policy that is nearly optimal after a small number of iterations.

\begin{figure}[ht]
\Ebox{1.0}{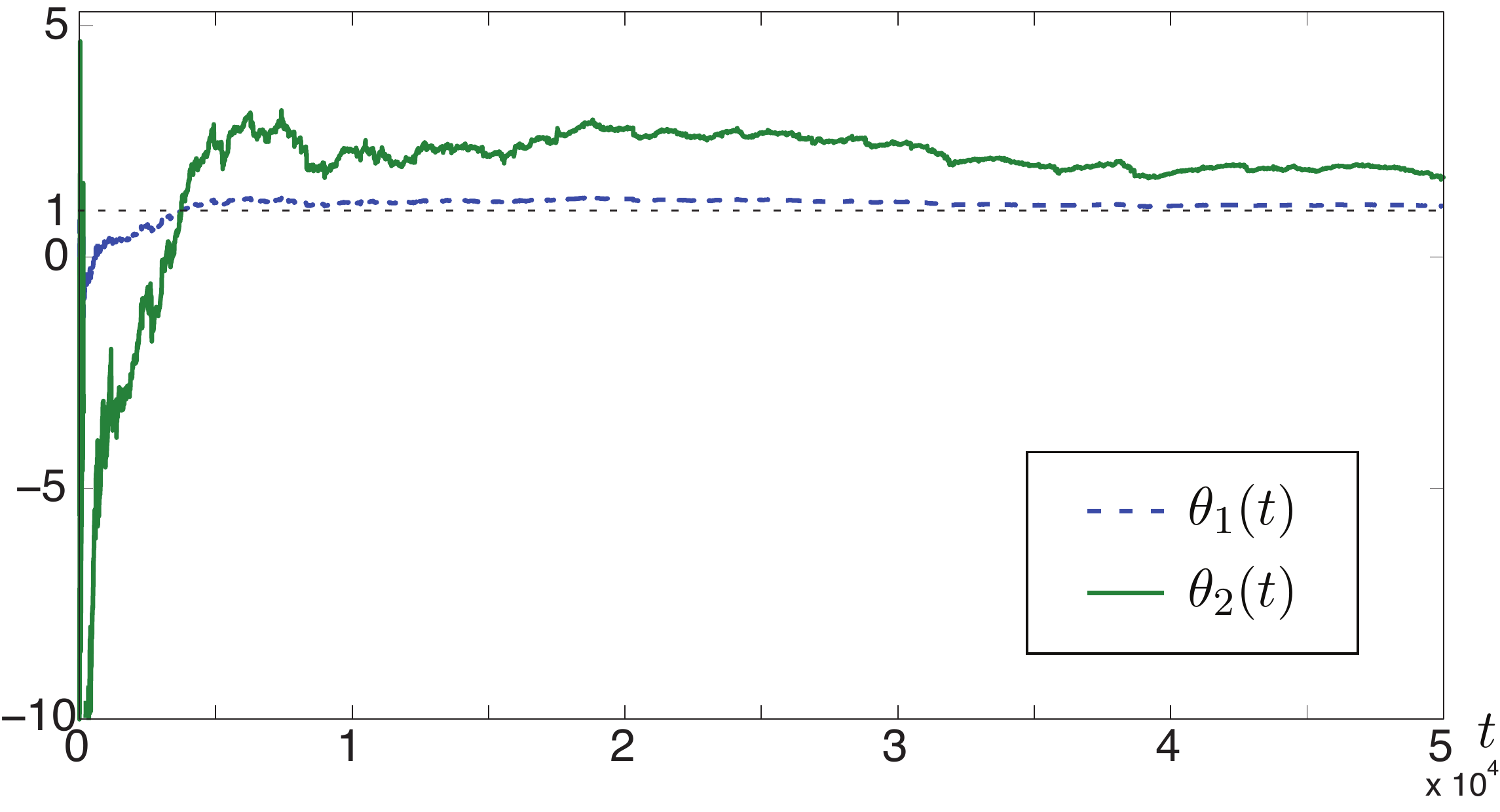}
     \caption{Convergence of the two parameters in TD-learning using the basis in \eqref{e:PolyBasis}.}
\label{fig:quadraticcosttheta}
\end{figure}

\begin{figure}[ht]
 \Ebox{1}{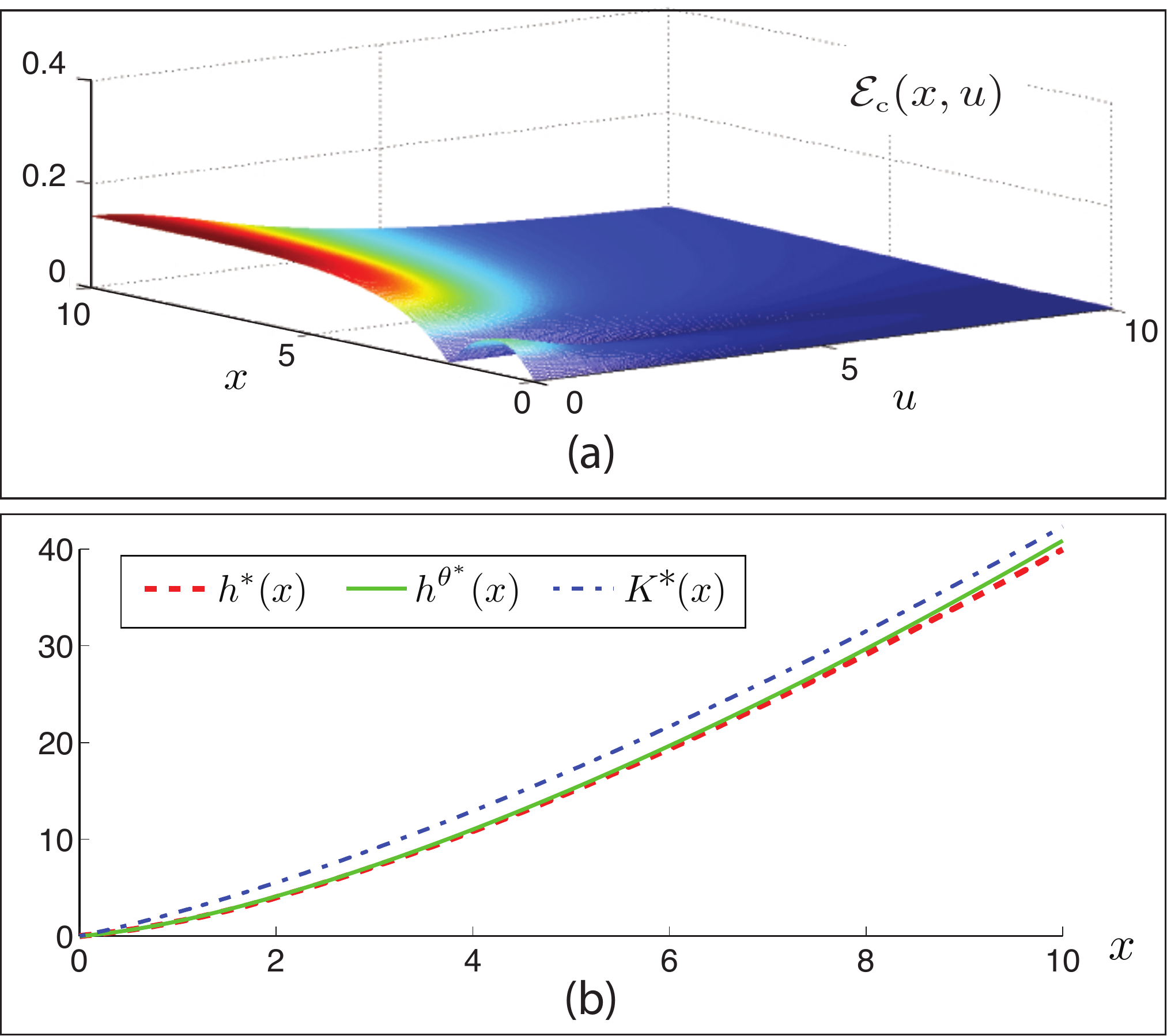}
\caption{Value functions and normalized error. The samples of arrival process are generated according to the scaled geometric distribution given in \eqref{e:A1}.   (a) The     normalized error $\Ecost$.   (b)  Comparison of the final approximation $h^{\theta^*}$, the fluid value function $K^*$, and the relative value function $h^*$.}
\label{f:quadraticcostvalue1}
\end{figure}

\begin{figure}[ht] 
\Ebox{1}{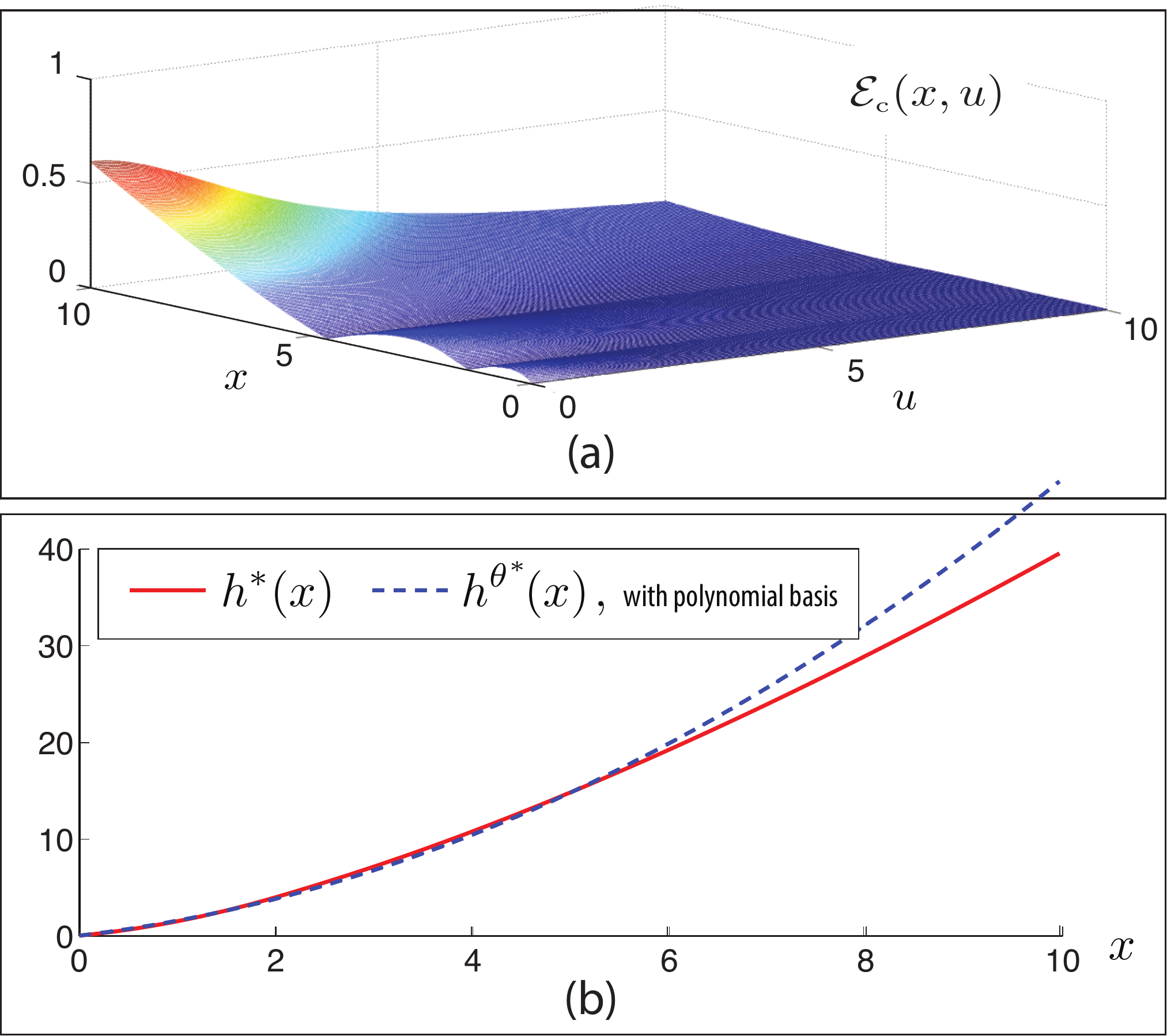}
\caption{Value functions and normalized error for the polynomial basis \eqref{e:PolyBasiscomp}.   (a) The     normalized error $\Ecost$.   (b) The comparison of the final approximation $h^{\theta^*}$ and the relative value function $h^*$.}
\label{f:quadraticcostvalue2}
\end{figure}

\Figure{fig:quadraticcosttheta} shows  the estimates of the coefficients obtained  after $50,000$ iterations of the LSTD algorithm,    after four steps of policy iterations.  The value of the optimal coefficient $\theta_1^*$ corresponding to $\psi_1=K^*$ was found to be close to unity, which is consistent with  \eqref{e:diffusionbasis}.

Shown in
  \Figure{f:quadraticcostvalue1} are error plots for the final approximation of $h^*$ from TDPIA.    \Figure{f:quadraticcostvalue1}~(a) is the Bellman error -- note that it is less than unity for all $(x,u)$.
  \Figure{f:quadraticcostvalue1}~(b) provides a comparison of two approximations to the solution to the relative value function $h^*$.    It is clear that the approximation $h^{\theta^*}$ obtained from the TDPIA algorithm     closely approximates the relative value function.

To illustrate the results obtained using a generic polynomial basis,  identical experiments were run using 
\begin{equation}
\psi'_1(x) = x,\quad \psi'_2(x) =  x^2     ,\qquad x\ge 0.
\label{e:PolyBasiscomp}
\end{equation}
The experiments were \textit{coupled}, in the sense that
the sample path of the arrival process was held fixed in  experiments comparing the results from the two different basis sets.

  \Figure{f:quadraticcostvalue2} shows error plots for the final approximation of $h^*$ using this quadratic basis for TD-learning.  In (a) we see that the Bellman error is significantly larger for $x> 5$, when compared with the previous experiments using the basis \eqref{e:PolyBasis}.  Similarly, the plots shown in \Figure{f:quadraticcostvalue1}~(b) show that the quadratic basis does not give a good fit for $x> 5$.

\subsection{The impact of variability}
 \label{s:Arho}

The influence of variability was explored by running the TDPIA with arrival distributions of increasing variance, and with mean fixed to unity.

The variance $\sigma^2_A$ is denoted $\kappa$, which is taken as a variable.  The specification of the marginal distribution is described informally as follows:  A weighted coin is flipped.  If a head is obtained, then $A_\kappa(t)$ is a scaled Bernoulli random variable.  If a tail, then $A_\kappa(t)$ is a realization of the scaled geometric random variable defined in \eqref{e:A1}.   Thus, this random variable can be expressed,
\begin{equation}
A_\kappa(t) =(1-B(t)) A_0(t)+B(t) \Delta_Z Z(t)  ,\qquad t\ge1,
\label{e:A2}
\end{equation}
where $A_0(t)$, $B(t)$, and $Z(t)$ are mutually independent,
$B(t)$ is a Bernoulli random variable with parameter $\varrho$,   $Z(t)$ is a Bernoulli random variable with parameter $\varrho_Z$,
and $ \Delta_Z=1/\varrho_Z$.  Hence the mean of $A_\kappa$ is unity:
\[
\alpha_\kappa=(1-\varrho)\alpha_0+\varrho \varrho_Z\Delta_Z  =1.
\]
The variance of $A_\kappa$ is a function of the parameters $\varrho$ and $\varrho_Z$:
\begin{equation}
\sigma^2_{A_\kappa}=  (1-\varrho)\sigma^2_{A_0}+\varrho\sigma^2_{Z}{\Delta_Z}^2= (1-\varrho)+\varrho(1-\varrho_Z)/\varrho_Z.\nonumber
\end{equation}
The parameter $\varrho$ was chosen to be $0.9$,  so that for a given $\kappa = \sigma^2_{A_\kappa}$ we obtain,
\[
\varrho_Z = 9/(8+10\kappa)
\]

Eight values of $\kappa$ were considered.  To reduce the relative variance,
the simulations were coupled as follows:  At each time $t$, the  eight  Bernoulli random variables were generated as follows:
\begin{equation}
\begin{aligned}
Z_1(t)&=B_1(t),\\
Z_{i+1}(t)&=Z_i(t)+(1-Z_i(t))B_{i+1}(t), \quad 1\leq i \leq 7.\nonumber
\end{aligned}
\end{equation}
where $B_i(t)\,,  1\leq i\leq 8$ are mutually independent Bernoulli random variables with parameters $\varrho_i$.  The arrival processes $\bfmA_\kappa$ were then generated by using $\bfmB$ and $\bfmZ$ according to \eqref{e:A2}.    The parameters $\varrho_i$ were chosen such that the eight variances  were $1$, $2$, $4$, $8$, $12$, $16$, $24$ and $32$.

The TDPIA was run 1000 times in parallel for these eight systems.  Four steps of policy improvement were performed:   In each step of policy iteration, the value function was estimated with $30,000$ iterations of the LSTD algorithm.  The final estimation of the coefficients from TDPIA were projected onto the interval $[-20, 20]$.

\begin{figure}[ht]
\Ebox{.75}{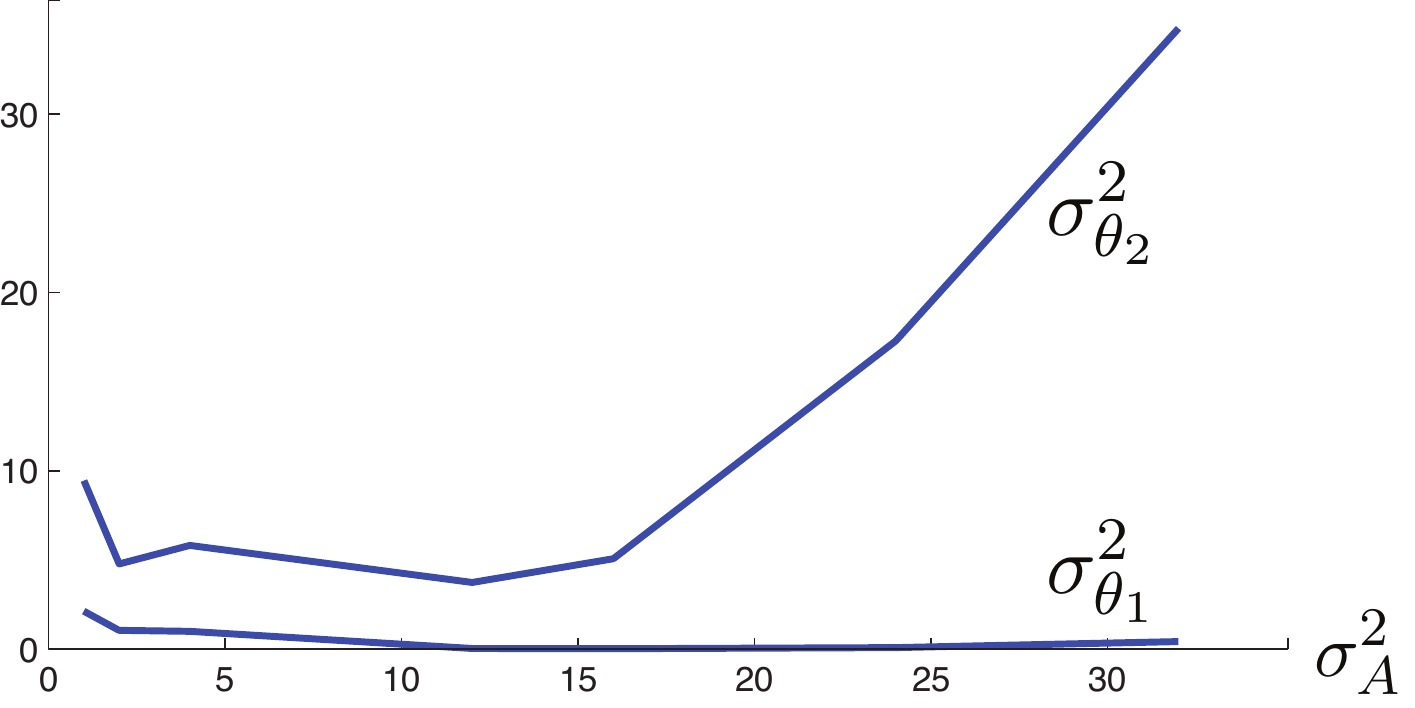}
\caption{The empirical variance obtain from 1000 independent experiments using   TDPIA. The variance of $\theta_2$ is far larger than $\theta_1$.  }
\label{f:varpara}
\end{figure}

\begin{figure}[ht]
\Ebox{1.0}{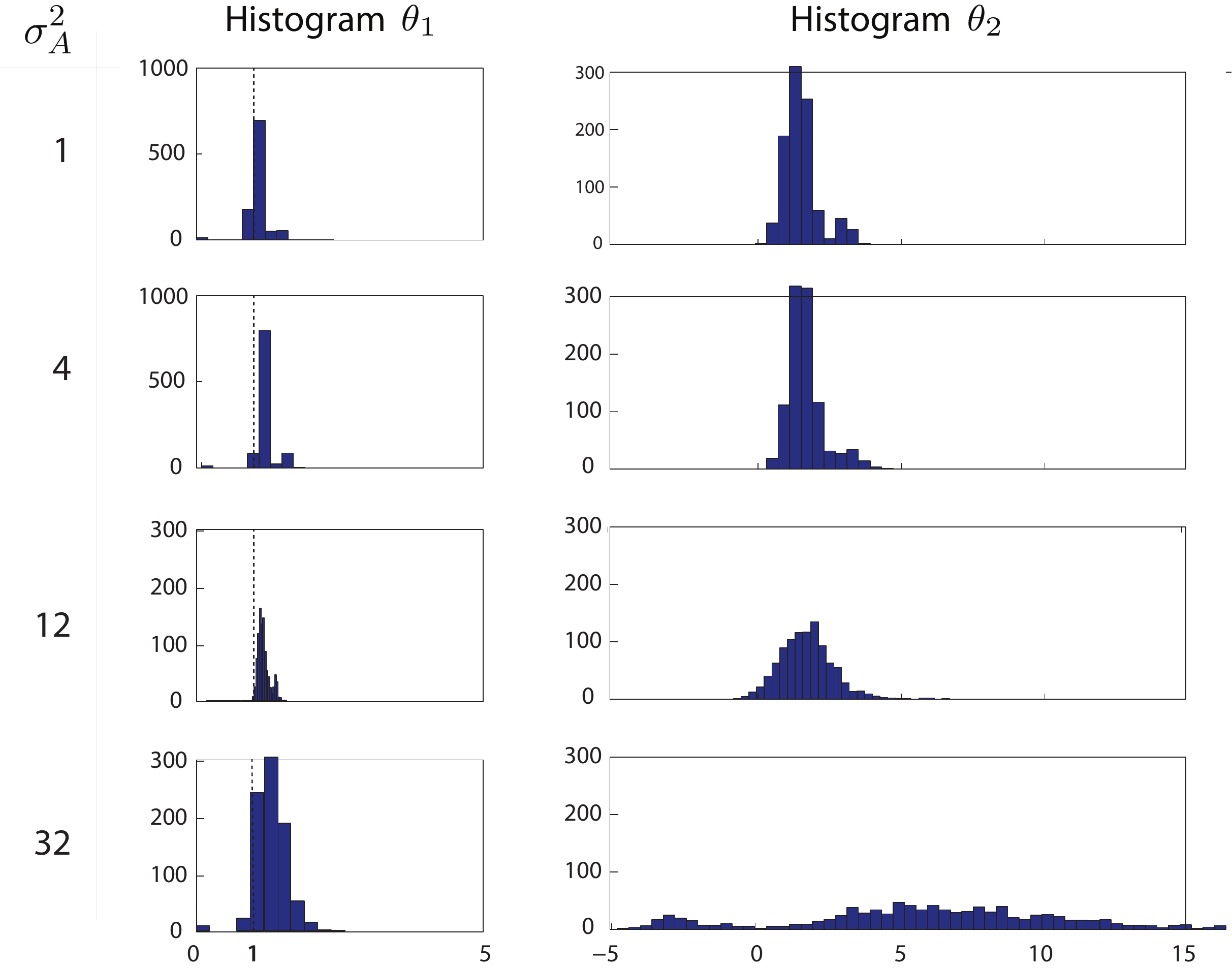}
\caption{Histograms of coefficients from 1000 simulations of TDPIA  with arrival processes generated according to the parametric family of distributions given in \eqref{e:A2}.   }
\label{f:comphist}
\end{figure}

\Figure{f:varpara} shows that the empirical variance of the final estimates of the coefficients. The variance for $\theta_2$ is much
larger than $\theta_1$ and increases with the variance of the arrival process when the variance is larger than $12$. This is also indicated by the histograms of the final estimations of the coefficients given in \Figure{f:comphist}.  

\begin{figure}[ht]
 \Ebox{1}{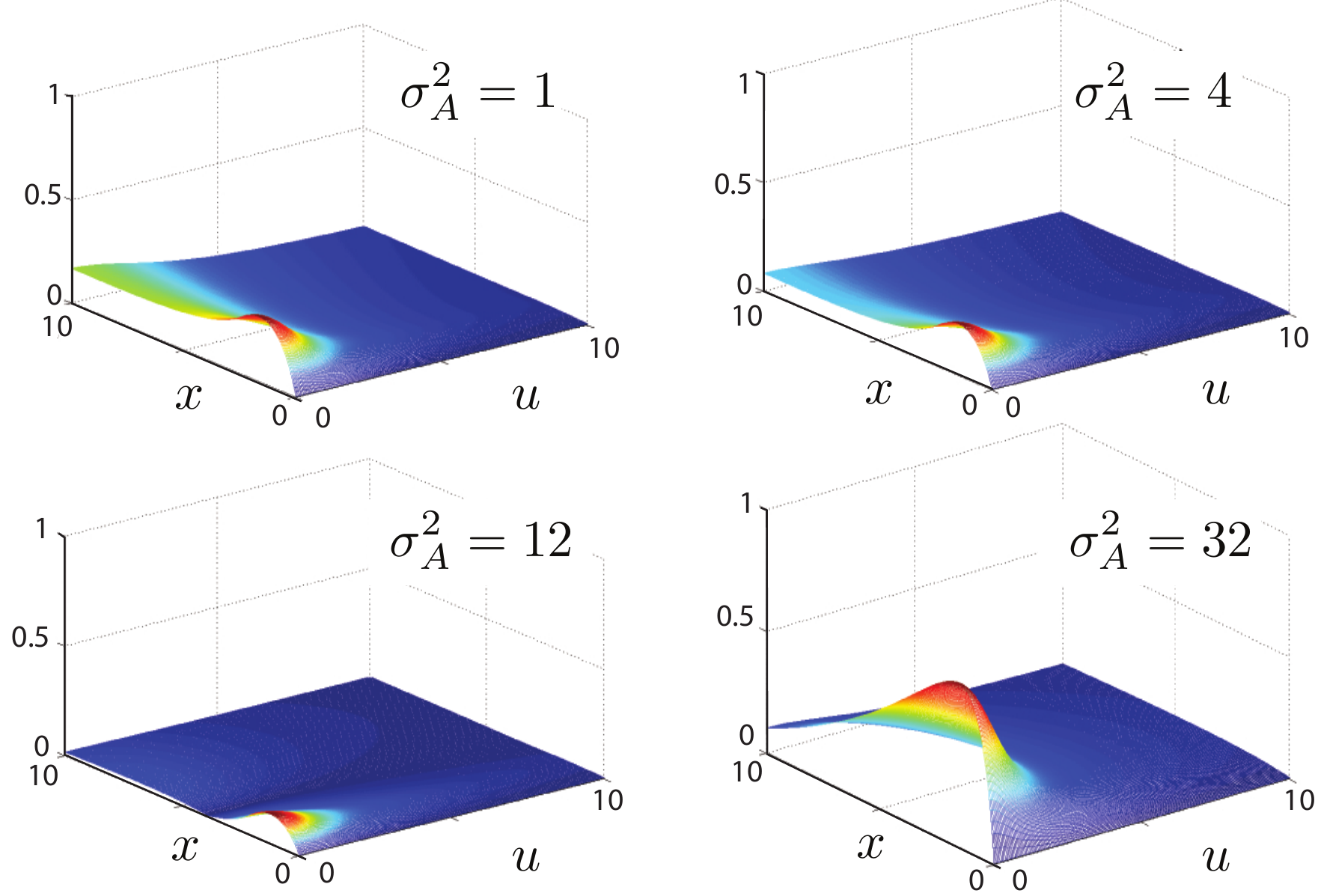}
\caption{Comparison of normalized error with arrival processes generated according to the parametric family of distributions given in \eqref{e:A2}.}
\label{f:comperror}
\end{figure}

Let $\bar{\theta}$ denote the mean of the $1000$ final estimations of coefficients.  A comparison of normalized Bellman errors for $h^{\bar{\theta}}$ is given  in \Figure{f:comperror}. The normalized error of $h^{\bar{\theta}}$ is largest when the variance of the arrival process is $32$.

\section{Concluding remarks}
\label{s.conclusion}

The main message of this paper is that idealized models
are useful for designing the function class for TD-learning.
This approach is applicable for control synthesis and performance approximation of Markov models in a wide range of applications.   
Strong motivation for this approach is provided by a Taylor series arguments that can be used to bound the difference between the relative value function $h^*$ and 
approximations based on fluid or diffusion models.

We have focused on the problem of power management in processors via dynamic speed scaling in order to illustrate the application of this approach for TD-learning. This application reveals that this approach to approximation yields remarkably accurate results.  Bounds for the Bellman error and the direct error w.r.t.\ the fluid and diffusion approximations indicate that the value functions of these idealized models are good approximations of the solution to the ACOE.
In particular, numerical experiments revealed that  value iteration initialized using the fluid approximation results in much faster convergence, and policy iteration coupled with TD-learning quickly converges to an approximately optimal policy when   the fluid and diffusion models are considered in the construction of a   basis. Besides, by using the fluid and diffusion value functions as basis functions, the Bellman error for the value function from TDPIA is much smaller than that obtained using quadratic basis functions.



\section*{Acknowledgment}
Financial support from the National Science Foundation (ECS-0523620 and CCF-0830511),
AFOSR   FA9550-09-1-0190,
ITMANET  DARPA RK 2006-07284, European Research Council (ERC-247006) and Microsoft Research is gratefully acknowledged.\footnote{Any opinions, findings, and conclusions or
  recommendations expressed in this material are those of the authors
  and do not necessarily reflect the views of NSF, AFOSR, DARPA,  or Microsoft.
  }

\clearpage

\appendix 

\noindent{\large\bf List of symbols and definitions}

\begin{tabular}{p{0.8 cm} p{6.90cm}}
$X(t)$ & State at time $t$; values in $\state$.
\\
$U(t)$ & Action at time $t$;  values in $\U$.
\\
$P_u$ & Transition kernel of the MDP, \eqref{e:Ptran}.
\\
$c$ & Cost function.
\\
$\eta^*$ &Minimum average cost.
\\
$\tau_{x}$ & The first return time to state  $x\in\state$.
\\
$\Edir$ & Direct error, \eqref{e:omega}.
\\
$\Ebell$ & Bellman error for the MDP model, \eqref{e:Belltheta}.
\\
$\Ebell^\tD$  & Bellman error for the diffusion model, \eqref{e:Bellh}.
\\
$\Ecost$ & Normalized error, \eqref{e:Bcr}.
\\
$h^*$ & Relative value function.
\\
$J^*$ & Fluid value function. 
\\
$\phi^{h}$ & The $(c,h)$-myopic policy, \eqref{e:myopic}.
\\
$\phi^*$ & The $(c,h^*)$-myopic policy, \eqref{e:ACOEF}.
\\
$\phiF$ & Fluid optimal policy, \eqref{e:TCOEF}.
\end{tabular}

\begin{tabular}{p{0.8 cm} p{6.90cm}}
$\generate_u$ & Generator in discrete time, \eqref{e:Dgen}.
\\
$\generateF_u$ & First-order ordinary differential operator, \eqref{e:GenFa}.
\\
$ \generateD_u$ & Second-order ordinary differential operator, \eqref{e:diffusionGen}.
\\
$\clP(u)$ & Power consumption term in cost function \eqref{e:cP}.
\end{tabular}

\section{Fluid value function and its properties}
\label{s:fluidandprop}

For computation it is simplest to work with modified cost functions, defined as follows 
\begin{equation}
\begin{aligned}
\hbox{\it Polynomial cost}\ \ &c^{\tF}(x,u) = x+  \nu ( [u- \alpha]_+)^\varrho,
\\
\hbox{\it Exponential cost}\ \ &c^{\tF}(x,u) = x+  \nu [e^{\kappa  u} -e^{\kappa  \alpha}]_+,
\end{aligned}
\label{e:PSnormCost}
\end{equation}
where $[\varble]_+ = \max(0,\varble)$.
The corresponding fluid value functions  are given in the following.

Part~(i) of \Proposition{convexJ} exposes a connection between the
fluid control policy and prior results on worst-case algorithms for speed scaling \cite{BKP07}.

\begin{proposition}
\label{t:convexJ}
The fluid value functions for the speed scaling can be computed or approximated for general $\alpha$:
\begin{romannum}
\item 
For  polynomial cost,  the value function and optimal policy for the fluid model are given by,
\begin{eqnarray}
J^*(x) &=& \nu x^{\frac{2\varrho - 1}{\varrho}}
    \frac{\varrho^2}{2\varrho - 1}\Bigl(\frac{1}{\nu(\varrho -1)}\Bigr)^{\frac{\varrho-1}{\varrho}}
\label{e:polyJ}
\\[.25cm]
\phiF(x) &=&
\Bigl(\frac{x}{\nu(\varrho -1)}\Bigr)^{1/\varrho} 
+\alpha. 
\label{e:polyJphi}
\end{eqnarray}

\item For   exponential cost, the fluid value function satisfies the following upper and lower bounds:
On setting  $\tilbeta=\nu e^{\kappa \alpha}$ and $\tilx =
x-\tilbeta $, there are constants $C_-,C_+$ such that the following holds whenever  $x
\geq \tilbeta (e^2 +1)$,
\[
C_-
+\frac{\kappa}{2}\frac{\tilx^2}{\log (\tilx)-\log(\nu)-({\kappa
\alpha+1})}\leq J^*(x) \leq C_+ +\frac{\kappa}{2}\tilx^2.
\]

\end{romannum}
\end{proposition}

\proof 
(i)
From \eqref{e:hfirst1}, the value function for the fluid model solves the following
equation:
\begin{equation}
\label{e:fluid}
0=\min_{u\geq 0}\bigl(x+  \nu ( [u- \alpha]_+)^\varrho+\nabla
J^*(x)\cdot(-u+\alpha)\bigr).
\end{equation}
Equivalently,
$$ x+  \nu ( [\phiF(x)- \alpha]_+)^\varrho = \nabla
J^*(x)\cdot(\phiF(x)-\alpha),$$
where $u = \phiF$ is the minimizer of the function $u \mapsto x + \nu ( [u- \alpha]_+)^\varrho+\nabla
J^*(x)\cdot(-u+\alpha)$. For each $x$, this function is decreasing on $[0,\alpha]$; thus $\phiF(x) \geq \alpha$.
Using the first-order optimality condition gives
\begin{equation}
\phiF(x) =\bigr(\frac{1}{\nu\varrho}\nabla J^*(x)\bigr)^{\frac{1}{\varrho-1}}+\alpha.
\label{e:fluidutemp}
\end{equation}
By substituting \eqref{e:fluidutemp} into \eqref{e:fluid}, we have
$x=(\nu\varrho-\nu)(\phiF(x)-\alpha)^{\varrho}$, which gives the desired value function and optimal policy.

\smallbreak

We now prove (ii).
The fluid value function satisfies the dynamic programming equation
\[
0=\min_{u\geq 0}\bigl( x+  \nu [e^{\kappa  u} -e^{\kappa  \alpha}]_++\nabla J^*(x)\cdot(-u+\alpha)\bigr).
\]
Arguing as in case (i), we see that the minimizer of the right-hand-side $\phiF(x)$ satisfies $\phiF(x) \geq \alpha$.
The first-order optimality condition gives:
\begin{equation}
\label{e:opte}
\kappa \nu e^{\kappa \phiF(x)}=\nabla J^*(x).
\end{equation}
By substituting \eqref{e:opte} for $\nabla J^*(x)$ into the dynamic programming equation,
and after some manipulation, we arrive at
\[x-\tilde{\nu}=e\tilde{\nu} e^{\kappa
(\phiF(x)-\alpha)-1}[\kappa(\phiF(x)-\alpha)-1],\]
where $\tilde{\nu}=\nu e^{\kappa\alpha}$.
Letting $w=\kappa(\phiF(x)-\alpha)-1$, we can write the above
as,
\begin{equation}
\label{e:opte1}
\frac{x-\tilde{\nu}}{e\tilde{\nu}}=e^ww,
\end{equation}
alternatively
\begin{equation}
\label{e:ue}
w=W\left( \frac{x-\tilde{\nu}}{e\tilde{\nu}}\right)  \quad {\rm and} \quad
\phiF(x)=\frac{W(\frac{x-\tilde{\nu}}{e\tilde{\nu}})+1}{\kappa}+\alpha,
\end{equation}
where $W$ is the Lambert W function.

To show the bounds on $J^*$, write $J^*(x)=C_1+\int_{b}^{x}\nabla J^*(s)ds$.
Now
$$\nabla J^*(s) = \kappa e\tilde{\nu} e^{W\left( \frac{s-\tilde{\nu}}{e\tilde{\nu}} \right)} = \kappa e \tilde{\nu}\frac{\frac{s-\tilde{\nu}}{e\tilde{\nu}}}{W\left( \frac{s-\tilde{\nu}}{e\tilde{\nu}} \right)} \geq ke \tilde{\nu}
\frac{\frac{s-\tilde{\nu}}{e\tilde{\nu}}}{\log\left( \frac{s-\tilde{\nu}}{e\tilde{\nu}} \right)},$$
where the inequality holds for $\frac{s-\tilde{\nu}}{e\tilde{\nu}} \geq e$. Using the substitution
$t=\frac{s-\tilde{\nu}}{e\tilde{\nu}}$, taking
$b=\tilde{\nu}(e^2+1)$ and letting $y = \frac{x-\tilde{\nu}}{e\tilde{\nu}}$ we obtain
\begin{eqnarray}
J^*(x)&\geq &C_1+\kappa (e\tilde{\nu})^2 \int_{e}^y \frac{t}{\log
t}dt\nonumber\\
&\geq & C_1+\kappa (e\tilde{\nu})^2 \frac{t^2}{2\log
t}\big|_{e}^y\nonumber\\
&=&C_{-}+\frac{\kappa }{2} \frac{\tilde{x}^2}{\log (\tilde{x})-\log(\nu)-({\kappa \alpha+1})}.\nonumber
\end{eqnarray}
Again using that $W(y)\leq \log y$, we can derive an upper bound on $J^*$ as follows,
\begin{eqnarray}
J^*(x)&=&C_2+\int_{e}^{y}\kappa e\tilde{\nu} e^{W(t)} e\tilde{\nu}dt\nonumber\\
&\leq &C_2+\kappa e\tilde{\nu} e\tilde{\nu}\int_{e}^y tdt\nonumber\\
&=&C_{+}+\frac{\kappa}{2} {\tilde{x}^2}.\nonumber
\end{eqnarray}
This completes the proof. 
\qed

 The next section applies elementary calculus to obtain bounds on value functions.

\section{Error bounds}
\label{s:Error bounds}
We now establish  the bounds on direct and Bellman error surveyed in the paper.  
The first proof establishes   bounds on the direct error in terms of the Bellman error:

\paragraph{Proof of \Proposition{errorboundh}}
We first obtain an upper bound on the direct error $\Edir(x)$. Based on the assumptions, $h^*\, (x)$ and $h\, (x)$ have the SSP representations. Consequently, the direct error can be written as,
\begin{equation}
\begin{aligned}
&h^*\, (x)-h\, (x)\\
=&
\min_{\bfmU}\Expect_x [\sum_{t=0}^{\tau_{x^\circ}-1}(c(X(t),U(t))-\eta^*)]
 \\
&-
 \Expect_x^{\phi^{h}}[\sum_{t=0}^{\tau_{x^\circ}-1}(c^h_{\phi^{h}}(X(t)) - \eta)]
 \\
 \leq&
\Expect_x^{\phi^{h}}[\sum_{t=0}^{\tau_{x^\circ}-1}(c_{\phi^{h}}(X(t))-\eta^*)]
\\
&-
 \Expect_x^{\phi^{h}}[\sum_{t=0}^{\tau_{x^\circ}-1}(c^h_{\phi^{h}}(X(t)) - \eta)]
\\
 =&
\Expect_x^{\phi^{h}}[\sum_{t=0}^{\tau_{x^\circ}-1}(\Ebell(X(t))-\eta^*)],\nonumber
\end{aligned}
\end{equation}
where the last equality follows from the definition of the perturbed cost in \eqref{e:ctheta}.

The proof of the lower bound is identical.
\qed

Bounds on the Bellman error are obtained using elementary calculus in the following.

\paragraph{Proof of \Proposition{clEbdds}}
By the Mean Value Theorem, there exists a random variable  $\barX_l$ between $x$ and $x+f(x,\phiJstar(x),W(1))$ such that,
\[
\begin{aligned}
\generate_{\phiJstar}J^*(x)
 =&
\Expect\bigl[ J^*(X(t+1)) - J^*(X(t)) \mid \\
&
 X(t)=x\, ,  \   U(t) = \phiJstar(x) \bigl]
\\
=&
 \Expect [
 \nabla J^*(x)^\transpose\cdot f(x,\phiJstar(x),W(1))\\
&+\half  f(x,\phiJstar(x),W(1))^\transpose \cdot \nabla^2J^*\,  (\barX_l)\\
&\quad \cdot  f(x,\phiJstar(x),W(1))].
\end{aligned}
\]

From the definition of the
Bellman error  \eqref{e:Belltheta}, we have
\begin{align*}
\Ebell(x)
 =& \min_{u\in \U(x)}\bigl(c(x,u)  + \generate_u J^*\, (x)\bigr) \\
=& c_{\phiJstar}(x)+\generate_{\phiJstar}J^*(x)
\\ \nonumber
=&  \bigl[ c_{\phiJstar}(x) - \eta + \nabla J^*(x)^\transpose\cdot \barf(x,\phiJstar(x))\bigr] \\
&+ \half  \Expect [f(x,\phiJstar(x),W(1))^\transpose \cdot \nabla^2J^*\,  (\barX_l)
 \\
&\qquad \quad\cdot  f(x,\phiJstar(x),W(1))] + \eta.
\nonumber
\end{align*}
The first term in brackets is made smaller if $\phiJstar$ is replaced by $\phiF$,  and then vanishes by \eqref{e:hfirst1}.  This gives the desired lower bound on the Bellman error.
\medskip

The proof of the upper bound is similar:
From the definition of the Bellman error \eqref{e:Belltheta}, we have
\begin{equation}
\begin{aligned}
\Ebell(x) =& \min_{u\in \U(x)}\left(c(x,u)  + \generate_u J^*\, (x)\right) \\
\leq&   c_{\phiF}(x) +\generate_{\phiF(x)}J^*(x).
\label{e:ubtwoa}
\end{aligned}
\end{equation}

By the Mean Value Theorem, there exists a random variable  $\barX_u$ between $x$ and $x + f(x,\phiF,W(1))$ such that
\begin{equation}
\begin{aligned}
&\generate_{\phiF}J^*(x)\\
=&\Expect [  \nabla J^*(x)^\transpose \cdot f(x,\phiF(x),W(1))
\\
&
+\half   f(x,\phiF(x),W(1))^\transpose \cdot \nabla^2J^*\, (\barX_u)\cdot  f(x,\phiF(x),W(1))]. \nonumber
\end{aligned}
\end{equation}

Consequently,
\begin{equation}
\begin{aligned}
&c_{\phiF}(x)+\generate_{\phiF(x)}J^*(x)\\
= &\left[c_{\phiF}(x) -\eta + \nabla J^*(x)^\transpose\cdot \barf(x,\phiF(x)) \right]
 \\
&+\half  \Expect [ f(x,\phiF(x),W(1))^\transpose \cdot \nabla^2J^*\, (\barX_u)
\\
&\qquad \quad
\cdot  f(x,\phiF(x),W(1))] + \eta
\end{aligned}
\label{e:clEupper1}
\end{equation}
The first term vanishes by \eqref{e:hfirst1}.

Combining \eqref{e:ubtwoa} and \eqref{e:clEupper1} gives the desired upper bound on the Bellman error.
\qed

The following properties of the fluid value function with $\clP$ quadratic  
will be used repeatedly in the analysis that follows.  We note that part (i) of the lemma holds whenever $\clP$ is convex and increasing.  
\begin{lemma}
\label{t:fluidprop}
The fluid value functions $K^*$ in \eqref{e:KstarQuadA} and $J^*$ in \eqref{e:quadJ} share the following properties:
\begin{romannum}
\item
They are convex and increasing.
\item
Their first derivatives are concave and increasing.
\item
Their second derivatives are decreasing, and vanish as $x^{-\half}$ as $x\to\infty$.
\end{romannum}
\end{lemma}
\proof
The first derivative of $K^*$ in \eqref{e:KstarQuadA} is
\begin{equation}
\nabla K^*(x)=\alpha+(2x+\alpha^2)^{\half }.
\label{e:nablaK}
\end{equation}
 It is an increasing, concave and positive function.  Consequently, the function $K^*$ is convex and increasing.

The second derivative of $K^*$ is
\begin{equation}
\nabla^2 K^*(x) = (2x+\alpha^2)^{-\half }.
\label{e:secondK}
\end{equation}
It  is positive, decreasing, and vanishes as $x^{-\half}$ as $x\to\infty$.

The proof for the properties of $J^*$ is identical.
\qed

\paragraph{Proof of \Proposition{DiffBellman0}}

The convexity and monotonicity of $K^*$ is established in \Lemma{fluidprop}.

The Bellman error $\Ebell^\tD$ for $K^*$ is
\begin{equation}
\begin{aligned}
\Ebell^\tD(x) & = \min_{u\geq 0}\bigl(c(x,u)+\generateD_u K^*(x)\bigr)\\
& = \min_{u\geq 0}\bigl(c(x,u)+(-u+\alpha) \nabla K^*(x) \bigr) + \half  \sigma_A^2 \nabla ^2 K^*(x)
\end{aligned}
\end{equation}
The minimum  vanishes by \eqref{e:hfirst1}, which implies the desired bound on $\Ebell^\tD(x)$.
\qed

\paragraph{Proof of \Proposition{DiffBellman}}
It is easy to verify that the derivative of $h$ defined in \eqref{e:diffusionbasis} is increasing, so that $h$ is convex.
Moreover,  it is non-decreasing since  $\nabla h(0) \ge 0$.


Using the first-order optimality condition in the definition of the Bellman error gives the minimizer in \eqref{e:Bellh},
\begin{equation}
\phi(x)= \nabla h\, (x) .
\label{e:diffpoli}
\end{equation}
Substituting  \eqref{e:generatordiff} and \eqref{e:diffpoli} into \eqref{e:Bellh} gives,
\begin{equation}
\begin{aligned}
\Ebell^\tD(x)=&\eta(2x+\alpha^2)^{\half }(2x+q^2)^{-\half }-\half \eta^2 (2x+q^2)^{-1}
\\
&+\half\sigma_A^2\bigl((2x+\alpha^2)^{-\half}+\eta(2x+q^2)^{-\frac{3}{2}}\bigr).
\label{e:bellh1}
\end{aligned}
\end{equation}
The first term in $\Ebell^\tD$ can be approximating using Taylor series,
\begin{equation}
\begin{aligned}
&\eta(2x+\alpha^2)^{\half }(2x+q^2)^{-\half }
\\
=&
\eta(1+\frac{\alpha^2-q^2}{2}(2x+q^2)^{-1}+\mathcal{O}(x^{-2})),
\label{e:bellh2}
\end{aligned}
\end{equation}
and combining \eqref{e:bellh1} and \eqref{e:bellh2} gives the desired conclusion.
\qed

To compute the  bounds on $\Ebell$ in \eqref{e:lowerKbell}, we first give some properties of the   $(c,K^*)$-myopic policy $\phiKstar$ in the following lemma.
 \begin{lemma}
\label{t:phiKstarprop}
 The policy $\phiKstar$ satisfies  
  \[
\lim_{x\rightarrow \infty} \frac{\phiKstar (x)  }{\sqrt{x}}  = \frac{1}{\sqrt{2}}.
\] 
\end{lemma}
\proof
On denoting $\overline{K}^*(x) =\Expect[K^*(x+A(1))]$, we can express the $(c,K^*)$-myopic policy by using the first-order optimality condition  in the definition \eqref{e:myopic},
\begin{equation}
\phiKstar(x)=\nabla\overline{K}^*(x-\phiKstar(x)). \nonumber
\end{equation}
Since $\nabla K^*(x)$ is a concave function of $x$, a result given in \Lemma{fluidprop}, 
by Jensen's inequality we obtain
\begin{equation}\label{e:ineqclPuJ}
\begin{aligned}
\nabla\overline{K}^*(x-\phiKstar(x))=& \Expect[\nabla K^*(x  - \phiKstar(x) + A(1))]
\\
\leq&  \Expect[ \nabla K^*(x+A(1))]
\\
\leq& \nabla K^*(x+\Expect[A(1)])
\\
=&\nabla K^*(x+\alpha)\, .
\end{aligned}
\end{equation}
The first equality is justified because $A(1)$ has bounded support and $K^*$ is $C^2$, so we can justify the exchange   of expectation and derivative.   
The first inequality follows from convexity of $K^*$.

Equations \eqref{e:ineqclPuJ} combined with \eqref{e:nablaK} imply the upper bound, 
\begin{equation}
 \phiKstar(x)=\mathcal{O}(x^{\half }).
\label{e:corela1}
\end{equation}
We next establish a limit.

Since $\nabla K^*$ is increasing, as shown in \eqref{e:nablaK}, we obtain
\begin{equation}
 \phiKstar(x)=\nabla\overline{K}^*(x-\phiKstar(x)) \geq \nabla K^*(x- \phiKstar(x)).
\label{e:phiKl}
\end{equation} 
Combining \eqref{e:ineqclPuJ} and \eqref{e:phiKl} gives,
\begin{equation}
\label{e:phiKstarlimit}
\lim_{x\rightarrow \infty}  \phiKstar (x)  = \lim_{x\rightarrow \infty} \nabla K^*(x).
\end{equation}
Substituting  \eqref{e:nablaK} into \eqref{e:phiKstarlimit} then gives the desired conclusion.
\qed

\paragraph{Proof of \Proposition{clEpos} Part (i)}
The lower bound of $\Ebell$ follows directly from \Proposition{clEbdds}~Part~(i),
\begin{equation}
\begin{aligned}
\Ebell(x) &\geq \half  \Expect \left[ \nabla^2K^*(\barX_l)\cdot (- \phiKstar(x)+A(1))^2 \right]+\eta
 \\
&\geq  \half  \Expect \left[ \nabla^2K^*(x+A(1))\cdot (- \phiKstar(x)+A(1))^2 \right] \geq 0,
\label{e:lowerKbell}
\end{aligned}
\end{equation} 
where the second and third inequalities follow from \Lemma{fluidprop} that  the second derivative of $K^*$ 
 is positive and decreasing.

Combining   \Lemma{phiKstarprop} and Assumption~\As{1} we obtain \begin{equation}
\begin{aligned}
&\lim_{x\rightarrow \infty}  \frac{1}{\sqrt{2x}}   \Expect \left[ \nabla^2K^*(x+A(1))\cdot (- \phiKstar(x)+A(1))^2 \right]      = 1
\label{e:Belllowerlim}
\end{aligned}
  \end{equation}

Similarly, the upper bound of the Bellman error follows from \Proposition{clEbdds}~Part~(ii),
\begin{equation*}
\begin{aligned}
\Ebell(x) &\leq \half   \Expect \left[ \nabla^2K^*(\barX_u)\cdot (-\phiF(x)+A(1))^2\right]
\\
 &\leq \half   \Expect \left[ \nabla^2K^*(x-\phiF(x))\cdot (-\phiF(x)+A(1))^2\right],\nonumber
\end{aligned}
\end{equation*}
where 
the second inequality follows from the fact that $\nabla^2K^*$ is decreasing.

The fluid optimal policy based on $K^*$ is $\phiF(x)=\nabla K^*(x)$,
and \eqref{e:secondK}  provides an expression for the second derivative of $K^*$.     
Substituting  these expressions in the previous limit gives,
\begin{equation}
\begin{aligned}
&\lim_{x\rightarrow \infty}  \frac{1}{\sqrt{2x}}    \Expect \left[ \nabla^2K^*(x-\phiF(x))\cdot (-\phiF(x)+A(1))^2\right]  =1
\label{e:Bellupperlim}
\end{aligned}
\end{equation}
Combining \eqref{e:Belllowerlim} and \eqref{e:Bellupperlim} gives the desired conclusion.
\qed

To bound the direct error,  we begin with a lemma useful to obtain finer bounds on value functions.
A drift condition for this purpose is a relaxation of \eqref{e:fish} called \textit{Poisson's inequality} \cite{chemey99a,CTCN}:
For some policy $\phi$,  a constant $\bar\eta<\infty$,  and a function $V\colon\Re_+\to\Re_+$,
\begin{equation}
c(x) + \generate  V\, (x) \le  \bar\eta, \quad x\ge 0.
\label{e:fishineqcm}
\end{equation}
where $c(x) = c_{\phi}(x) = c(x,\phi(x))$,  and  $\generate= \generate_{\phi}$, as in \eqref{e:fish}.

It is simplest to state the following result for a general Markov chain on $\Re_+$:
\begin{proposition}
\label{t:rootV3}
Suppose that   Poisson's inequality \eqref{e:fishineqcm} holds,  and that the continuous functions $V$ and $c$ satisfy for some $1\le z_c<z_v$, and $\epsy>0$,
\[
V(x) \le\epsy^{-1} x^{z_v},\quad c(x) \ge \epsy x^{z_c},\qquad  x\ge 1.
\]
Then there exists a bounded set $S$ such that for any scalar $\rho$ satisfying $\rho<1$ and $\rho \ge (1-z_c/z_v)$,   there exists a solution to (V3) of \cite{MT},
\[
PV_\rho \le V_\rho - c_\rho + b\ind_S
\]
where $S$ is a bounded set,  $b$ is a constant,  and the functions satisfy, for perhaps a different $\epsy>0$,
\[
V_\rho(x) \le\epsy^{-1} x^{\rho z_v},\quad c_\rho(x) \ge \epsy x^{z_c - (1 -\rho) z_v},\qquad  x\ge 1.
\]
\end{proposition}

\proof
The function $V_\rho$ is given by $V_\rho=  (1+V)^\rho$.  
First note that Poisson's inequality implies a version of (V3),
\[
PV \le V - c_1 + b_1\ind_{S_1}
\]
where $c_1=1+\half  c$, $b_1$ is a constant, and $S_1$ is a bounded interval.   This holds because of the assumptions on $c$.

 The bound is then a simple application of concavity:
\begin{equation}
\begin{aligned}
V_\rho(X(t+1)) \le& V_\rho(X(t)) + \rho (1+V(X(t)))^{\rho-1} \\
&\bigl( V(X(t+1)) - V(X(t))\bigr)
\end{aligned}
\end{equation}
Taking conditional expectations given $X(t)=x$ gives,
\[
\begin{aligned}
PV_\rho\, (x) &\le V_\rho(x) + \rho (1+V(x))^{\rho-1} \bigl( PV\, (x) - V(x)\bigr)
\\
&\le V_\rho(x) + \rho (1+V(x))^{\rho-1} \bigl( - c_1 (x)  + b_1\ind_{S_1}(x) \bigr)
\end{aligned}\]
which implies the result.
\qed

%

The following lemma is based on the Comparison Theorem in \cite{MT}.
\begin{lemma}
\label{t:stoppingtime}
The following inequality holds,
\[
\Expect_x^{\phiKstar}[\sum_{t=0}^{\tau_0-1}\Ebell(X(t))]\leq b_1 x+b_2,
\]
where $b_1$ and $b_2$ are two constants. 
\end{lemma}
\proof
By \Proposition{inve}, the fluid value function $K^*$ also satisfies the ACOE,
\begin{equation}
\min_{u\in\U(x)}\bigl(c^{K^*}(x,u)+ \generate_u K^*\,(x) \bigr)  = \eta,
\label{e:ACOEQinv}
\end{equation}
where $c^{K^*}$ is the inverse dynamic programming solution  given in \eqref{e:ctheta}, and $\eta\in \Re_+$ is an arbitrary constant.

It shows that the following Poisson's equation holds,
\[
P V=V-c+\eta,
\]
where  $P=P_{\phiKstar}$, $c=c^{K^*}(x,{\phiKstar}(x))$, and $V=K^*$ is a Lyapunov function.
Applying \eqref{e:KstarQuadA} and \eqref{e:Ecostlim}, 
there exist $x_0>0$ and $1>\epsy>0$ such that
\[
V(x) \le\epsy^{-1} x^{\frac{3}{2}},\quad c(x) \ge \epsy x,\qquad  x\ge x_0.
\]

Consequently,  \Proposition{rootV3} implies that there exists a solution of (V3):
\[
PV_\rho \le V_\rho - c_\rho + b\ind_S,
\]
where $\rho=2/3$, $S$ is a bounded set,  $b$ is a constant,  and the functions satisfy, for $\epsy_1>0$,
\[
V_\rho(x) \le\epsy_1^{-1} x,\quad c_\rho(x) \ge \epsy_1 x^{\half},\qquad  x\ge x_0.
\]

Following \Proposition{clEpos} and Theorem~(14.2.3) in \cite{MT} gives,
\[
\Expect_x^{\phiKstar}[\sum_{t=0}^{\tau_0-1}\Ebell(X(t))]\leq a_1\Expect_x^{\phiKstar}[\sum_{t=0}^{\tau_0-1}\bigl(X(t)^{\half }+a_2\bigr)]\leq b_1 x+b_2,
\]
where $a_1$, $a_2$, $b_1$ and $b_2$ are constants.
\qed

\paragraph{Proof of \Proposition{clEpos} Part~(ii)}

We first obtain an upper bound on $\Edir$. Combining \Proposition{errorboundh} and  \Lemma{stoppingtime} gives the desired upper bound on the direct error:
\spm{Warning!  Check to see if $\eta^*$ makes sense here}
\begin{equation}
\begin{aligned}
\Edir(x)& \leq \Expect_x^{\phiKstar}[\sum_{t=0}^{\tau_0-1}(\Ebell(X(t))-\eta^*)] \\
&\leq \Expect_x^{\phiKstar}[\sum_{t=0}^{\tau_0-1}\Ebell(X(t))]  \leq b_1x +b_2. \nonumber
\end{aligned}
\end{equation}

The proof of the lower bound is similar.   \Proposition{clEpos} implies that there exist a constant $b>0$ and a bounded set $S$ such that,
\spm{Warning!  Check to see if $\eta^*$ makes sense here}
\begin{equation}
 \Ebell(x)-\eta^*\geq -b \ind_S.
\label{e:lowerbellS}
\end{equation}
By Lemma~11.3.10 in \cite{MT} and the bound \eqref{e:small} assumed in Assumption~\As{2}, we have $\Expect_x^{\phi^*}[\sum_{t=0}^{\tau_{0}-1} \ind_{S}]<n$, where $n>0$ is a constant.
Consequently, \Proposition{errorboundh}  together with \eqref{e:lowerbellS} gives the lower bound on the direct error,
\[
\Edir(x)\geq  \Expect_x^{\phi^*}[\sum_{t=0}^{\tau_{0}-1}(\Ebell(X(t))-\eta^*)]\geq -\Expect_x^{\phi^*}[\sum_{t=0}^{\tau_{0}-1} b\ind_{S}]\geq - b n.
\]
\qed

\bibliographystyle{plain}
\bibliography{inequality,power,strings,markov,q}

\end{document}